\documentclass[pdflatex,sn-mathphys-ay]{sn-jnl}

\usepackage{graphicx}
\usepackage{multirow}
\usepackage{amsmath,amssymb,amsfonts}
\usepackage{amsthm}
\usepackage{mathrsfs}
\usepackage{xcolor}
\usepackage{textcomp}
\usepackage{manyfoot}
\usepackage{booktabs}
\usepackage{algorithm}
\usepackage{algorithmicx}
\usepackage{algpseudocode}
\usepackage{listings}
\usepackage{tabularx}
\usepackage{array}
\usepackage[english]{babel}
\usepackage{enumitem}
\usepackage{cleveref}
\usepackage{subcaption}
\usepackage{overpic}

\newcolumntype{P}[1]{>{\raggedright\arraybackslash}p{#1}}

\theoremstyle{thmstyleone}

\theoremstyle{thmstyletwo}

\theoremstyle{thmstylethree}

\begin{document}

\title[Falling Behind Drives Unsafe Development]{Falling Behind Drives Unsafe Development in an Idealised AI Race Experiment}

\author*[1,2]{\fnm{Elias} \sur{Fern\'andez Domingos}}\email{elias.fernandez.domingos@vub.be}
\author[3,4]{\fnm{The Anh} \sur{Han}}\email{T.Han@tees.ac.uk}

\affil*[1]{\orgdiv{AI-Lab}, \orgname{Vrije Universiteit Brussel}, \orgaddress{\street{Pleinlaan 9}, \city{Brussels}, \postcode{1050}, \country{Belgium}}}
\affil[2]{\orgdiv{MLG}, \orgname{Universit\'e Libre de Bruxelles}, \orgaddress{\street{Boulevard du Triomphe}, \city{Brussels}, \postcode{1050}, \country{Belgium}}}
\affil[3]{\orgdiv{School of Computing, Engineering and Digital Technologies, Teesside University}, \orgname{Teesside University}, \orgaddress{\city{Middlesbrough}, \country{United Kingdom}}}
\affil[4]{\orgdiv{Center for Digital Innovation}, \orgname{Teesside University}, \orgaddress{\city{Middlesbrough}, \country{United Kingdom}}}

\abstract{
Technological races create tension between speed and safety: actors may gain by moving faster than competitors, even when risky development is harmful. This is prominent in debates about artificial intelligence (AI), where competitive pressure is often argued to incentivise riskier, less safety-conscious development. We study this using a framed behavioural experiment based on an idealised AI race, in which paired participants repeatedly chose between Safe and Unsafe development under an uncertain time horizon. Unsafe development gave faster progress and higher immediate payoffs but accumulated private risk up to a treatment-specific maximum of 10\%, 60\%, or 90\%; the race's competitive structure was held constant, and only this maximum risk varied. Neither the pre-registered comparison between risk levels nor the role of elicited risk preferences was supported by the data. Instead, exploratory analyses motivated by the task's repeated structure show that Unsafe behaviour is shaped less by risk preferences than by the evolving strategic state of the race: participants are more likely to choose Unsafe after their opponent does so, being ahead reduces Unsafe play while falling behind increases it, and first-round choices predict later behaviour. To interpret these effects we introduce a reduced evolutionary model with four strategies -- Always Safe, Always Unsafe, Conditionally Safe, and Conditionally Antisocial Safe -- which reproduces the treatment effect and shows how conditional Unsafe behaviour can be favoured by competitive race dynamics. Together, the experiment and model show that unsafe development can emerge from early behavioural momentum, opponent behaviour, and fear of falling behind, rather than from risk preferences alone, suggesting policy should focus on reducing competitive pressure and promoting cooperation in AI development rather than only individual risk.
}


\maketitle

\section{Introduction}\label{sec:introduction}

Rapid technological development often takes place under competitive pressure. Firms, laboratories, or states may benefit from being first to develop, deploy, or control a new technology, but the same race incentives can also create pressure to accept higher levels of risk. This tension is especially salient for artificial intelligence (AI), where rapid progress in large language models (LLMs) and their widespread adoption have intensified concerns about whether competitive dynamics encourage actors to prioritise speed over safety \citep{hendrycks2023overview,ord2020precipice}. Recent cycles of advanced AI development, in which frontier AI labs and states compete to develop and deploy increasingly capable systems, have been characterised as AI (arm) races toward advanced or transformative AI \citep{bengio2024managing,gruetzemacher2025strategic,muller2026evolvable,saeri2026prioritization}.
Here, we ask how technological races shape human decisions when faster development is privately rewarding but potentially risky: when individuals face an AI race between Safe and Unsafe development, do they respond primarily to the level of risk, to their own risk preferences, or to the evolving competitive state of the race? To isolate this question, we hold the competitive structure of the race fixed -- two participants always compete under the same rules -- and instead manipulate the private cost of taking risks. As detailed below, the central finding is that Unsafe choices track the evolving state of this fixed competition rather than the level of risk or individual risk attitudes.

Economic theory has long studied races for innovation, market entry, and technological adoption. Classic models of patent races and pre-emption show that competition can accelerate innovation and induce actors to move early in order to secure strategic advantage \citep{Loury1979MarketStructureInnovation,FudenbergTirole1985Preemption}. However, while competition can stimulate innovation under some conditions, it can reduce incentives under others, as captured by the inverted-U relation between competition and innovation \citep{AghionEtAl2005CompetitionInnovation}. Additionally, in many real technological domains, speed is not merely a question of efficiency, but can trade off against risk mitigation.

In this direction, tournament and contest behavioural experiments show that participants often take greater risks when risk-taking improves relative performance, when the probability of winning is low, or when the contest is more winner-take-all \citep{BullSchotterWeigelt1987TournamentsPieceRates,NiekenSliwka2010RiskTakingTournaments,EriksenKvaloy2014MyopicRiskTaking,SchedlinskySommerWoehrmann2016RiskTakingTournaments,Spadoni2018CompetitionRiskTakingContests,GuertlerStruthThon2023CompetitionRiskTaking}. At the same time, behavioural evidence indicates that risk-taking under competition depends on institutional details, relative standing, gender, prior outcomes, and whether the individual has won or lost previous contests \citep{FilippinGioia2018CompetitionSubsequentRiskTaking,GenakosPagliero2012InterimRankRiskTaking}. Thus, the behavioural literature supports that competitive pressure can incentivise riskier behaviour, while also showing that the effect is context-dependent.

The AI-race literature makes the safety trade-off more explicit. Formal models of AI development races argue that first-mover advantages can make unsafe development individually attractive, even when unsafe actions increase the probability of harmful outcomes \citep{ArmstrongBostromShulman2016RacingPrecipice,han2020toRegulateOrNot}. Related work emphasises that race rhetoric and strategic competition may undermine cooperation over responsible AI development \citep{CaveOHeigeartaigh2018AIRace,AskellBrundageHadfield2019CooperationResponsibleAI,BarnettScher2025AIGovernanceExtinction}. Evolutionary and game-theoretical analyses of idealised AI races further show how incentives, regulation, heterogeneity, and institutional interventions can alter the prevalence of unsafe behaviour \citep{HanEtAl2021MediatingAI,han2022voluntary,CimpeanuEtAl2022AIRacesHeterogeneous}. More recent models of international technology races similarly analyse how uncertainty, information, and strategic competition can generate incentives for risky deployment or insufficient safety investment \citep{StaffordTragerDafoe2022SafetyNotGuaranteed,EmeryXuParkTrager2024TechnologyRaces,Polborn2025CompetitiveDevelopmentDangerousTechnologies,BuenoDeMesquitaDziudaPolborn2026AGIRace}. Across these theoretical contributions, competitive pressure is often argued to incentivise faster, riskier, or less safety-conscious technological development.

\begin{figure}[t]
    \centering
    \includegraphics[width=1.0\linewidth]{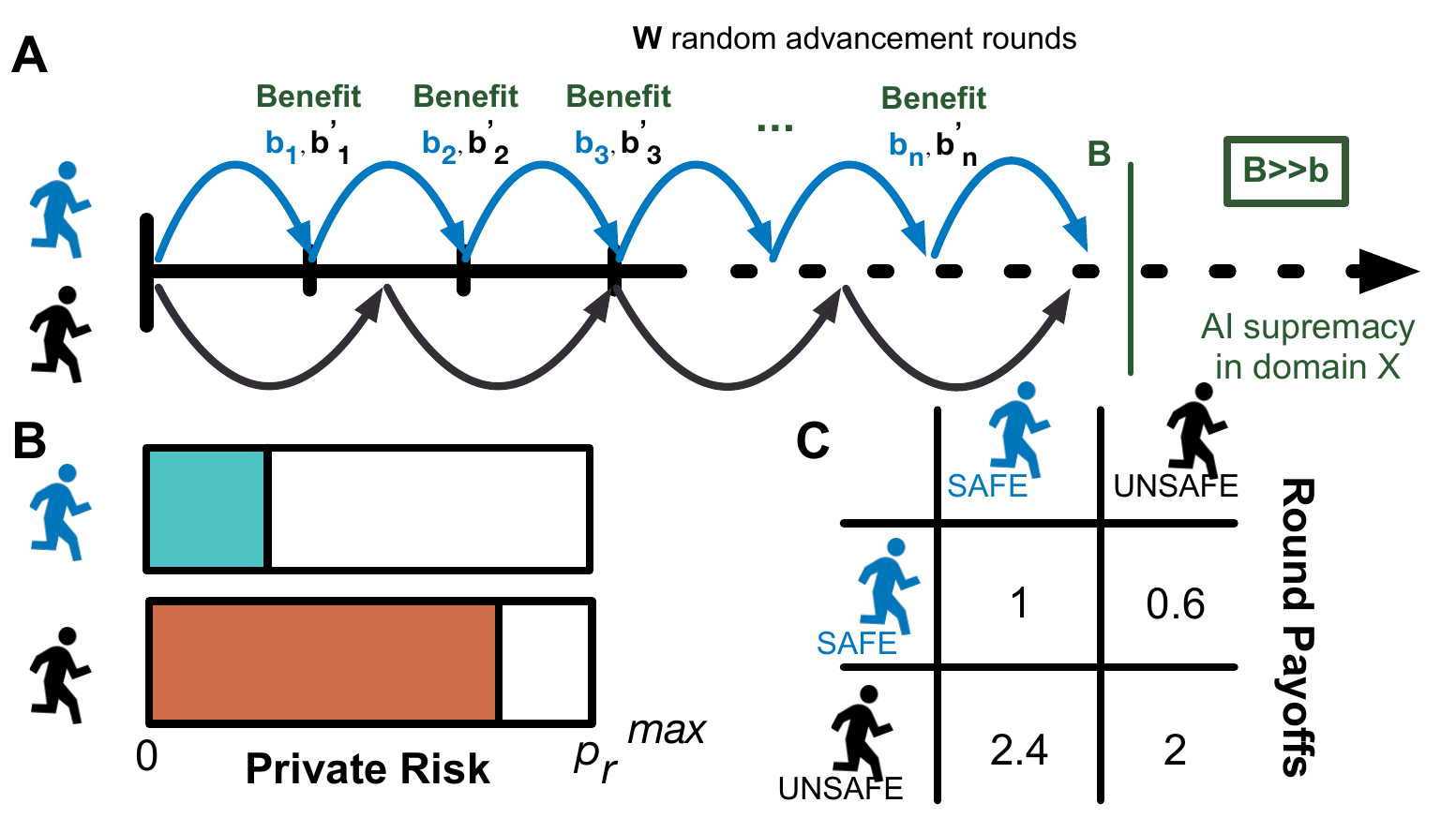}
    \caption{
    \textbf{Experimental AI race task.}
    (A) Participants were matched in pairs and repeatedly chose between Safe and Unsafe technological development under an uncertain time horizon. The game lasted for at least five rounds; after that, each additional round ended the race with constant probability. At each round, Safe development advances the participant position in the race $s_S=1$ step, while Unsafe development advances the race $s_U=1.5$ steps.
    (B) Unsafe choices increased a participant's private setback risk. The accumulated risk was proportional to the fraction of Unsafe actions chosen and bounded above by the treatment-specific maximum risk \(p_r^{\max}\).
    (C) Payoff matrix at each round. In summary, Unsafe development provides higher immediate payoff and faster technological progress, whereas Safe development reduces exposure to private risk.
    }
    \label{fig:ai_race_description}
\end{figure}

Despite this theoretical motivation, direct empirical evidence remains limited. Existing behavioural studies of contests and tournaments establish important mechanisms linking competition and risk-taking, but they rarely study repeated technological-race environments in which risky actions both accelerate progress and accumulate endogenous private risk (though see \citep{Zizzo2002RacingWithUncertainty} for a repeated patent-race experiment under technological and dynamic uncertainty). This leaves open the question of whether unsafe technological development is primarily driven by individual risk preferences, by the exogenous level of risk, or by the evolving strategic state of the race.

We address this question through a framed behavioural experiment based on an idealised two-player AI race (see Figure \ref{fig:ai_race_description} for a schematic representation). Participants were matched in pairs and repeatedly chose between Safe and Unsafe development. Unsafe actions generated higher immediate payoffs and faster race progress, but increased the participant's private probability of suffering a setback if they won or tied the race. The game lasted for a minimum of five of rounds and then ended with a constant probability ($p=0.2$) after each additional round, creating uncertainty about the race horizon.

Our preregistered primary experimental question asks whether the maximum level of private risk affects the frequency of Safe and Unsafe technological development (see \Cref{si:sec:prereg} for all details and differences with pre-registration). We manipulate the maximum private risk across three treatments,
\[
p_r^{\max}\in\{0.1,0.6,0.9\},
\]
and test the hypothesis that maximum private risk changes the odds of acting Safe, with higher maximum risk expected to increase Safe behaviour and reduce Unsafe behaviour. A preregistered secondary question asks whether participants' elicited risk preferences predict Unsafe choices, and whether this relationship is stronger when maximum private risk is high. Finally, we examine an exploratory question motivated by the theoretical AI-race literature: whether the observed behaviour can be interpreted as a distribution over simple strategies related to those proposed in idealised AI-race models \citep{han2020toRegulateOrNot}.

As detailed in Section \ref{sec:results} (Results), neither preregistered hypothesis was supported: we find no significant difference in Unsafe behaviour between the two originally pre-registered risk levels ($p_r^{\max}=0.6$ and $0.9$), and elicited risk preferences do not significantly predict Unsafe choices. Motivated by these null results, and by the fact that the repeated structure of the task allows us to analyse how behaviour depends on the evolving state of the race, we examine, as an exploratory analysis, whether Unsafe choices are predicted by the opponent's previous action, the participant's own previous action, and the relative race position. These dynamic variables are important because competitive pressure is unlikely to operate only through a static treatment effect. For example, falling behind, observing an opponent take risks, or establishing an early pattern of Unsafe play may each change the perceived incentives to prioritise speed over safety.

Our results show that Unsafe behaviour is primarily driven by interaction history and competitive position, rather than by maximum risk or elicited risk preferences alone. Participants are more likely to choose Unsafe after their opponent chose Unsafe in the previous round. Race position also matters: being ahead tends to reduce Unsafe choices, whereas falling behind sustains incentives to choose Unsafe. Moreover, first-round Unsafe choices predict later Unsafe behaviour, suggesting that early actions capture a persistent behavioural tendency or strategic signal.

To interpret these patterns, we develop a reduced evolutionary model of the experimental game. The model uses four behavioural strategies: Always Safe, Always Unsafe, Conditionally Safe, and a conditionally responsive strategy that starts with Unsafe. This reduced strategy set captures the main behavioural trends observed in the experiment while remaining numerically tractable. The model reproduces the qualitative treatment-level pattern and shows how conditional Unsafe behaviour can be favoured by competitive race dynamics. This strategic behaviour however is often omitted in previous AI-race models \citep{han2020toRegulateOrNot,HanEtAl2021MediatingAI,CimpeanuEtAl2022AIRacesHeterogeneous}.

Together, the experiment and model suggest that unsafe technological development is not explained by risk preference alone. Instead, it emerges from the interaction between early behavioural momentum, responsiveness to the opponent, and fear of falling behind in a competitive race, highlighting how AI and technological race risks can arise from strategic dynamics even when individual actors are not intrinsically risk seeking.

\section{Results}\label{sec:results}

We first report the results for the two pre-registered hypotheses (\Cref{si:sec:prereg-hypotheses}). Contrary to Hypothesis 1.2, we find no significant difference in the frequency of Unsafe choices between the $p_r^{\max}=0.6$ and $p_r^{\max}=0.9$ treatments (independent-samples $t=-0.0101$, $p=1$, Bonferroni-corrected; \Cref{fig:treatment_differences}A). Consistent with Hypotheses 2.1--2.3 not being supported, elicited risk preferences do not significantly predict Unsafe choices, either in the pre-registered mixed-effects analysis (\Cref{si:sec:prereg-analyses}) or in the panel regressions reported below (\Cref{si:sec:data-analysis-risk}). The remainder of this section therefore reports exploratory analyses of how Unsafe choices depend on interaction history and the evolving state of the race, motivated by these null results and by the repeated structure of the experimental task.

\subsection*{Unsafe choices are shaped by race position}

\begin{figure}[t]
    \centering
    \includegraphics[width=1.0\linewidth]{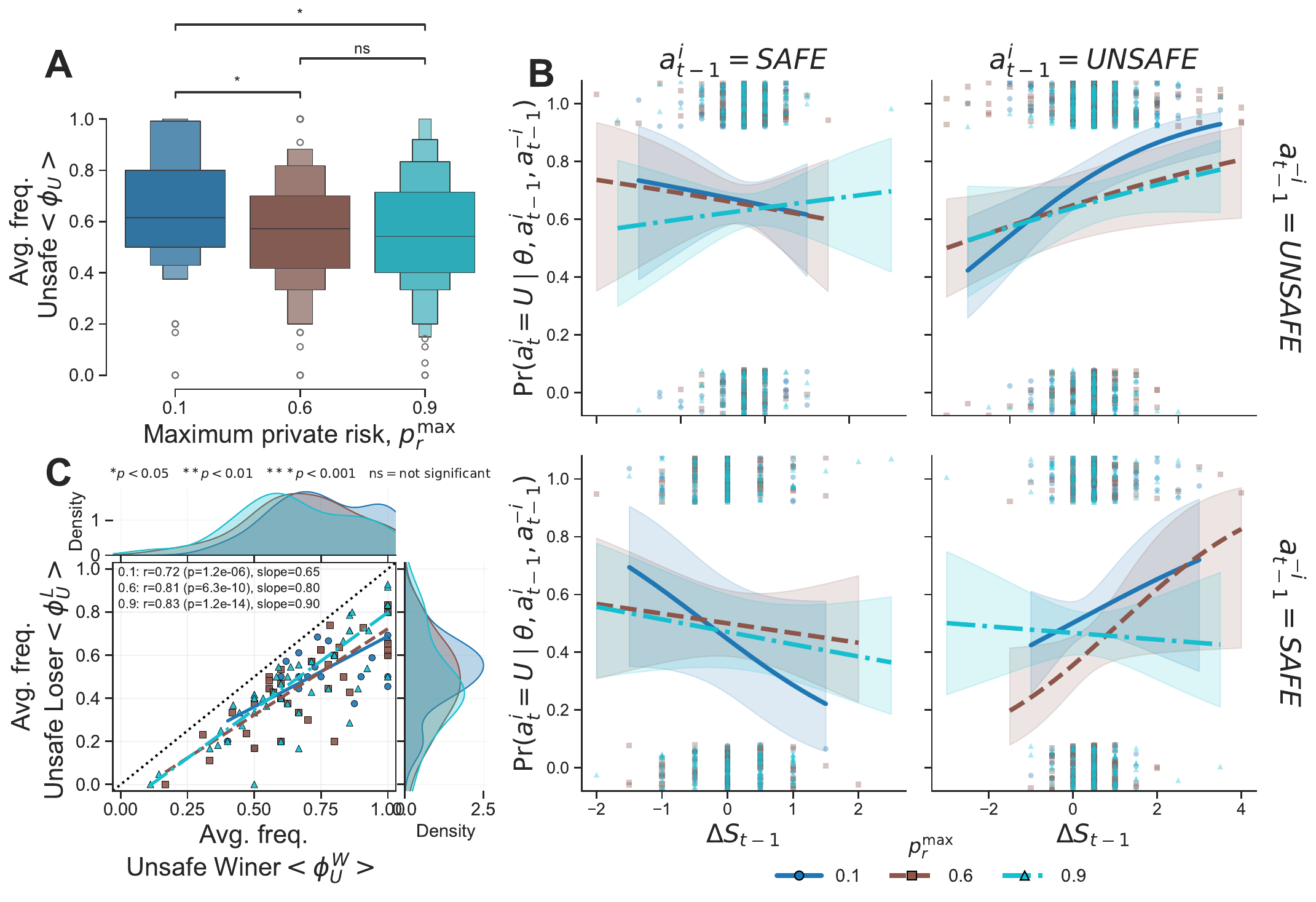}
    \caption{
        \textbf{Unsafe choices across treatments and race states.}
        (A) Average frequency of Unsafe choices across maximum private-risk treatments. Participants choose Unsafe frequently in all treatments, including at the highest private-risk level, $p_r^{\max}=0.9$, showing that the speed and payoff advantage of Unsafe development creates a persistent incentive to take risks. Pairwise independent-samples $t$-tests, Bonferroni-corrected across the three comparisons, show that Unsafe play is significantly higher in the $p_r^{\max}=0.6$ and $p_r^{\max}=0.9$ treatments than in the $p_r^{\max}=0.1$ treatment ($p_r^{\max}=0.1$ vs.\ $0.6$: $t=2.635$, $p=0.0272$; $p_r^{\max}=0.1$ vs.\ $0.9$: $t=2.811$, $p=0.0161$), with no significant difference between $p_r^{\max}=0.6$ and $p_r^{\max}=0.9$ ($t=-0.0101$, $p=1$). Tests are computed on each participant's mean Unsafe frequency across all rounds ($N=98$, $105$, $138$ for $p_r^{\max}=0.1$, $0.6$, $0.9$, respectively).
        (B) Unsafe choices as a function of the previous race position, $\Delta S_{t-1}$, the participant's previous decision, $a_i^{t-1}$, and the competitor's previous decision, $a_{-i}^{t-1}$. Participants who are behind in the race display a higher tendency to choose Unsafe than participants who are ahead, and participants are more likely to continue choosing Unsafe after mutual Unsafe play.
        (C) Relationship between the average frequency of Unsafe choices among race winners, $\langle \phi_U^W \rangle$, and race losers, $\langle \phi_U^L \rangle$. Winners tend to display higher average Unsafe frequencies than losers, but the two quantities are positively correlated within pairs. This correlation increases with $p_r^{\max}$, indicating that under higher maximum private risk, race outcomes are more strongly associated with the pair-level profile of Unsafe behaviour. The marginal distributions also show that the modal average frequency of Unsafe behaviour is lower when $p_r^{\max}=0.9$.
    }
    \label{fig:treatment_differences}
\end{figure}

We first examine how the frequency of Unsafe choices varies across maximum private-risk treatments and race states. \Cref{fig:treatment_differences} summarises the main experimental results. Participants choose Unsafe frequently across all treatments, including when $p_r^{\max}=0.9$, indicating that the speed and payoff advantage of Unsafe development creates a strong incentive to take risks (\Cref{fig:treatment_differences}A). However, Unsafe behaviour is not uniform across the interaction. It depends on the evolving state of the race, measured by the difference in race steps, $\Delta S$, as well as on the participant's previous decision, $a_i^{t-1}$, and the competitor's previous decision, $a_{-i}^{t-1}$ (\Cref{fig:treatment_differences}B). Participants who are behind in the race display a higher tendency to choose Unsafe than participants who are ahead. This is consistent with a fear-of-falling-behind mechanism: when a participant is losing, Unsafe development becomes attractive because it offers faster progress and a chance to recover. Conversely, when a participant is ahead, the incentive to continue taking risks is weaker, especially when maximum private risk is high. Additionally, participants are more likely to continue choosing Unsafe after mutual Unsafe play, whereas a previously Safe competitor increases the likelihood of switching to Safe behaviour (consistent with the cluster-robust logistic regression reported in \Cref{tab:cluster_logit_unsafe}). \Cref{fig:treatment_differences}C highlights this race-state dependence by comparing the average frequency of Unsafe choices among race winners, $\langle \phi_U^W \rangle$, and race losers, $\langle \phi_U^L \rangle$. Although winners tend to display higher average Unsafe frequencies than losers, both quantities are positively correlated, and this correlation increases with $p_r^{\max}$, suggesting that higher private risk makes race outcomes more strongly tied to the overall Unsafe profile of the pair. Additionally, the marginal distributions show that the modal average frequency of Unsafe behaviour is lower when $p_r^{\max}=0.9$.

These patterns suggest that participants do not respond only to the exogenous risk treatment. Instead, they condition their behaviour on the strategic state of the interaction. The remaining analyses therefore focus on previous behaviour, race-step differences, and first-round choices as predictors of later Unsafe development.

\subsection*{Opponent behaviour and first-round choices predict later Unsafe play}

\begin{table}[t]
\centering
\small
\caption{\textbf{Cluster-robust panel logistic regression of Unsafe decisions.} 
The dependent variable equals one if the focal player chose Unsafe in round \(t\). 
Standard errors clustered at the pair level are reported in parentheses. 
The reference treatment is \(p_r^{\max}=0.1\). 
\(\Delta S_{t-1}\) is centred around its sample mean. 
Models (1)--(3) exclude the initial-action bias \(a_i^1\), while models (4)--(6) include it. 
All models include demographic covariates and risk-preference covariates.
Demographic covariates control for participant sex, age, and nationality (South Africa or Poland, the two dominant recruitment pools, relative to all other nationalities), while risk-preference covariates are based on the \cite{eckel2008forecastingRiskAttitudes} risk-preference elicitation task.
Full estimates for these covariates are reported in \Cref{si:sec:data-analysis-demographics,si:sec:data-analysis-risk}.
\(^{\dagger}p<0.1\), \(^{*}p<0.05\), \(^{**}p<0.01\), \(^{***}p<0.001\).}
\label{tab:cluster_logit_unsafe}
\begin{tabular}{lcccccc}
\toprule
 & \multicolumn{3}{c}{Without \(a_i^1\)}  & \multicolumn{3}{c}{With \(a_i^1\)} \\
\cmidrule(lr){2-4}\cmidrule(lr){5-7}
 & (1) & (2) & (3) & (4) & (5) & (6) \\
\midrule
\(p_r^{\max}=0.6\) & -0.183 & -0.142 & -0.152 & -0.155 & -0.123 & -0.132 \\
 & (0.210) & (0.180) & (0.178) & (0.210) & (0.182) & (0.180) \\
\(p_r^{\max}=0.9\) & -0.245 & -0.190 & -0.194 & -0.241 & -0.188 & -0.191 \\
 & (0.183) & (0.155) & (0.155) & (0.181) & (0.155) & (0.155) \\
\(a_i^1\) &  &  &  & 0.291\(^{*}\) & 0.207\(^{\dagger}\) & 0.217\(^{\dagger}\) \\
 &  &  &  & (0.117) & (0.118) & (0.116) \\
\(a_i^{t-1}\) &  & 0.022 & -0.173 &  & 0.007 & -0.193 \\
 &  & (0.128) & (0.196) &  & (0.128) & (0.192) \\
\(a_{-i}^{t-1}\) &  & 0.863\(^{***}\) & 0.640\(^{**}\) &  & 0.832\(^{***}\) & 0.607\(^{**}\) \\
 &  & (0.123) & (0.197) &  & (0.120) & (0.192) \\
\(\Delta S_{t-1}\) &  & 0.106\(^{\dagger}\) & -0.238 &  & 0.065 & -0.296\(^{*}\) \\
 &  & (0.057) & (0.150) &  & (0.067) & (0.149) \\
\(a_i^{t-1}\times a_{-i}^{t-1}\) &  &  & 0.396\(^{\dagger}\) &  &  & 0.400\(^{\dagger}\) \\
 &  &  & (0.240) &  &  & (0.239) \\
\(a_i^{t-1}\times \Delta S_{t-1}\) &  &  & 0.448\(^{*}\) &  &  & 0.466\(^{*}\) \\
 &  &  & (0.185) &  &  & (0.182) \\
\(a_{-i}^{t-1}\times \Delta S_{t-1}\) &  &  & 0.201 &  &  & 0.218 \\
 &  &  & (0.213) &  &  & (0.214) \\
\(a_i^{t-1}\times a_{-i}^{t-1}\times \Delta S_{t-1}\) &  &  & -0.162 &  &  & -0.183 \\
 &  &  & (0.211) &  &  & (0.210) \\
Demographic covariates & Yes & Yes & Yes & Yes & Yes & Yes \\
Risk-preference covariates & Yes & Yes & Yes & Yes & Yes & Yes \\
Observations & 2,888 & 2,888 & 2,888 & 2,888 & 2,888 & 2,888 \\
Pair clusters & 172 & 172 & 172 & 172 & 172 & 172 \\
Pseudo \(R^2\) & 0.006 & 0.035 & 0.039 & 0.009 & 0.037 & 0.040 \\
\bottomrule
\end{tabular}
\end{table}

\Cref{tab:cluster_logit_unsafe} reports cluster-robust logistic regressions predicting whether participant $i$ chooses Unsafe in round $t$. The analysis uses decisions from round $2$ onward, giving $N=2{,}888$ observations across $338$ participants (see \Cref{si:sec:sample} for more information), with standard errors clustered at the pair level across $172$ pairs. The reference treatment is $p_r^{\max}=0.1$. Models (1)--(3) exclude the participant's first-round action, whereas models (4)--(6) include $a_i^1$, an indicator for whether participant $i$ chose Unsafe in round $1$.

All models include demographic and risk-preference covariates. Demographic covariates control for participant sex, age, and nationality, accounting for possible imbalances or behavioural differences associated with sample composition. Risk-preference covariates are based on the \cite{eckel2008forecastingRiskAttitudes} risk-preference elicitation task and were included because risk attitudes were part of the preregistered secondary hypotheses. To avoid cluttering the main table, the corresponding coefficients are reported in \Cref{si:sec:data-analysis-demographics,si:sec:data-analysis-risk}. These covariates do not significantly predict Unsafe choices and do not change the main effects reported below.

The most robust predictor of later Unsafe play is the opponent's previous action, which aligns with the patterns shown in \Cref{fig:treatment_differences}B. Participants are substantially more likely to choose Unsafe after observing the opponent choose Unsafe in the previous round. This effect remains positive and statistically significant across the additive and interaction specifications. In the full models, the coefficient of $a_{-i}^{t-1}$ is positive both without $a_i^1$ (model 3: $\hat{\beta}=0.640$, $p=0.001$) and with $a_i^1$ (model 6: $\hat{\beta}=0.607$, $p=0.002$). This indicates that Unsafe behaviour is not only a response to the exogenous risk treatment, but also to the competitor's recent behaviour.

By contrast, the participant's own previous action is not a robust predictor once the opponent's previous action and the race state are included. The coefficient of $a_i^{t-1}$ is small and statistically insignificant across specifications. This suggests that later Unsafe play is not explained by simple action persistence alone. Instead, participants respond strongly to the opponent's behaviour, which is consistent with conditional decision-making in the race.

Race position also matters. In the full specifications, the coefficient of $\Delta S_{t-1}$ is negative, indicating that being further ahead reduces the probability of choosing Unsafe when both players previously chose Safe. This effect is statistically significant when first-round action is included (model 6: $\hat{\beta}=-0.296$, $p=0.048$) but not significant without it (model 3: $\hat{\beta}=-0.238$, $p=0.113$). The positive interaction $a_i^{t-1}\times \Delta S_{t-1}$ shows that the effect of race position depends on the participant's previous action (model 3: $\hat{\beta}=0.448$, $p=0.016$; model 6: $\hat{\beta}=0.466$, $p=0.011$). In other words, the strategic meaning of being ahead or behind depends on the recent behavioural state of the interaction.

Including $a_i^1$ reveals that participants who chose Unsafe in round $1$ are more likely to choose Unsafe in later rounds. This association is significant in the baseline model with $a_i^1$ and remains marginally significant in the full interaction specification. First-round Unsafe choice therefore captures a persistent initial behavioural tendency that is not fully explained by subsequent conditional responses to previous-round decisions.

The direct treatment coefficients for $p_r^{\max}=0.6$ and $p_r^{\max}=0.9$ are negative relative to $p_r^{\max}=0.1$, but not statistically significant in the round $t\geq2$ panel regressions. This does not imply that maximum private risk is irrelevant. Rather, it suggests that after the first round, differences in Unsafe play are more strongly explained by interaction history and race position than by a simple direct treatment effect.

These results motivate the reduced strategy space used in the evolutionary model (\Cref{fig:model_predictions}). The strong effect of the opponent's previous action supports the inclusion of conditional strategies, and the effect of first-round Unsafe choice motivates distinguishing between conditional strategies that start Safe and those that start Unsafe. This leads to the four-strategy model analysed below. The relationship between this strategy set and the strategies studied in the repeated Prisoner's Dilemma literature is discussed in \Cref{si:sec:pd-relationship}.

\subsection*{A reduced evolutionary model reproduces the behavioural pattern}

\begin{figure}[h!]
    \centering
    \includegraphics[width=1\linewidth]{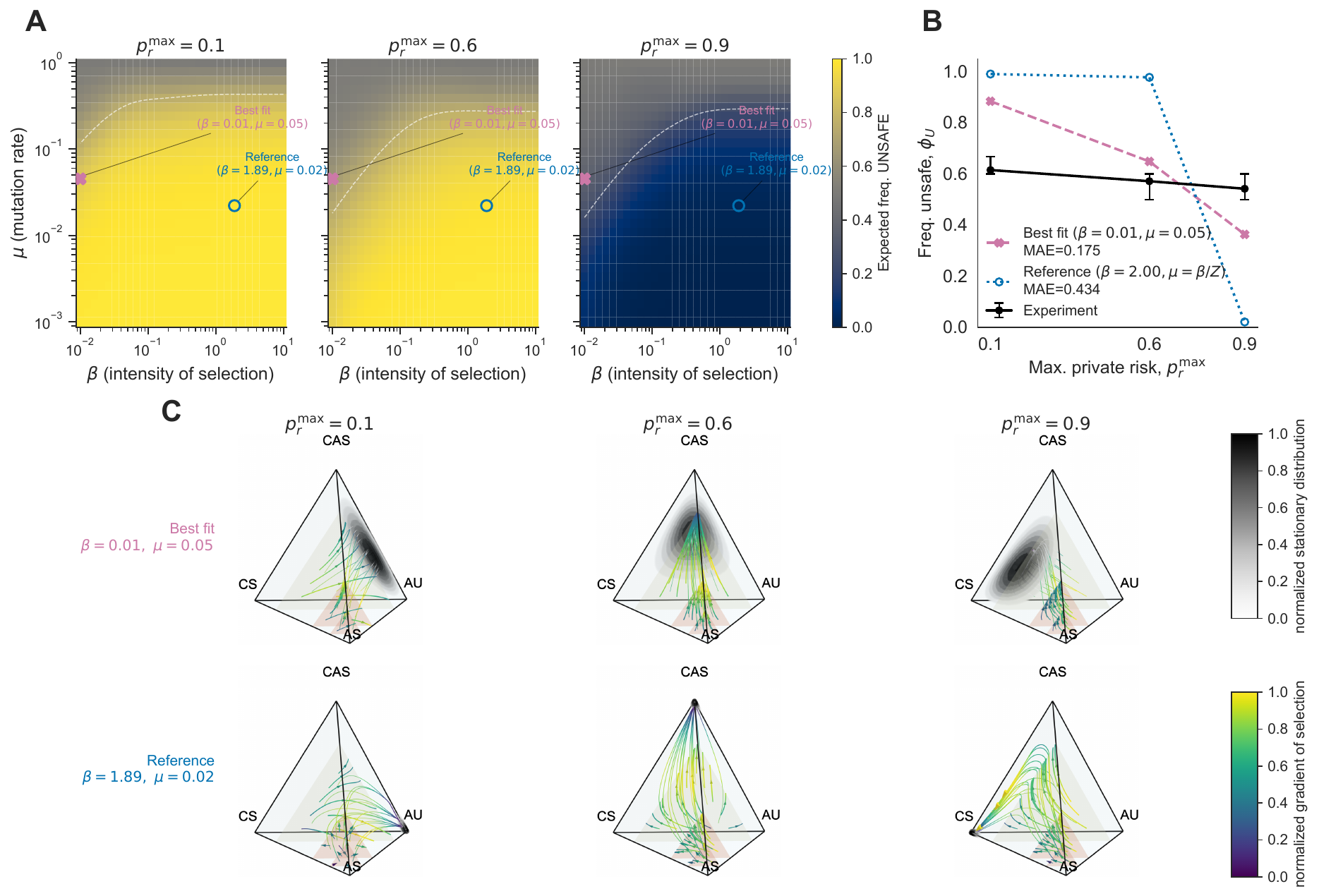}
    \caption{
    \textbf{Evolutionary model of Unsafe behaviour in the AI race.}
    (A) Model-predicted frequency of Unsafe choices as a function of the mutation rate, $\mu$, and the intensity of selection, $\beta$, across maximum private-risk treatments, $p_r^{\max}$. The blue circle marks a reference parametrisation with relatively strong selection, $\beta=2$, and mutation $\mu=\beta/Z$, slightly above neutral-drift. The second marked point shows the best-fitting parametrisation obtained when fixing a higher mutation rate, $\mu=0.05$, which captures the higher behavioural noise and implementation errors expected in the experiment.
    (B) Comparison between the experimental median frequency of Unsafe choices and the corresponding model predictions. The fitted model captures the qualitative direction of the experimental treatment effect, including the reduction in Unsafe behaviour at the highest private-risk level.
    (C) Simplex dynamics of the reduced four-strategy model with Always Safe (AS), Always Unsafe (AU), Conditionally Safe (CS), and Conditionally Antisocial Safe (CAS) strategies. Each point in the tetrahedral simplex represents a population state, with barycentric coordinates giving the frequencies of the four strategies. Arrows indicate the local direction of selection, and therefore the most likely evolutionary path from nearby states; arrow colour encodes the intensity of this selection gradient. Grey-scale spheres represent the finite-population stationary distribution over the three-dimensional simplex, with larger and darker spheres indicating states visited more frequently in the long run.
}
    \label{fig:model_predictions}
\end{figure}

The regression results suggest that participants condition their choices on the opponent's previous behaviour and on the state of the race. To study the game-theoretical consequences of such conditional behaviour, we construct a reduced evolutionary model of the experimental game (see \nameref{sec:methods} for more details). The model contains four strategies: Always Safe (AS), Always Unsafe (AU), Conditionally Safe (CS), and Conditionally Antisocial Safe (CAS). The two conditional strategies both respond to the opponent's previous action, but differ in their first-round action: CS starts with Safe, whereas CAS starts with Unsafe.

\Cref{fig:model_predictions} compares the predictions of this reduced model with the experimental data. \Cref{fig:model_predictions}A shows the model-predicted frequency of Unsafe behaviour across the relevant evolutionary parameter space. The model does not assume a separate behavioural rule for each risk treatment. Instead, treatment differences emerge from how maximum private risk changes the payoff consequences of Unsafe development, and therefore the evolutionary success of each strategy. We highlight two points in this parameter space. The reference point (blue circle) corresponds to relatively strong selection, $\beta=2$, and mutation $\mu=\beta/Z=0.02$, a standard choice representing selection strong enough to amplify payoff differences while keeping mutation just above the neutral-drift scale. By contrast, the best-fit point (pink cross) is obtained when fixing a higher mutation rate, $\mu=0.05$, and weak selection, $\beta=0.01$, which, in combination, capture a higher amount of behavioural noise expected in the experiment (see \Cref{si:sec:strategy-distribution} for a more in-depth analysis of the effect of $\beta$ and $\mu$ on the model).

\Cref{fig:model_predictions}B compares the experimental median frequency of Unsafe choices with the model predictions. The model captures the qualitative pattern across maximum private-risk treatments, including the relatively small difference between the two higher-risk treatments and the stronger contrast with the low-risk treatment. The best-fitting parameters therefore suggests that the experimental data are better described by a setting with more noise than the reference case, while still preserving the same broad strategic structure.

\Cref{fig:model_predictions}C shows the evolutionary dynamics in the simplex spanned by the reduced strategy set. These dynamics illustrate how conditional Unsafe behaviour can be maintained by the race structure. At the reference point, the stationary distribution is more concentrated and the dominant strategy depends strongly on $p_r^{\max}$. When private risk is low, Always Unsafe is favoured because Unsafe development provides a speed advantage at relatively low expected cost. At intermediate private risk, CAS becomes dominant, suggesting that beginning aggressively and then conditioning on the opponent's behaviour can be advantageous. At the highest private risk, CS dominates, indicating that Safe initial behaviour combined with conditional responses becomes more favourable when the accumulated cost of Unsafe development is large. Moving towards the best-fitting region increases mutation and weakens the effective force of selection, shifting probability mass away from the vertices and towards the faces and interior of the simplex. This yields a more diffuse stationary distribution, consistent with the greater heterogeneity observed in the experiment (see \Cref{si:sec:dynamics-3strategy} for more details on the effect of $p_r^{\max}$ on the dynamics). \Cref{si:sec:pd-relationship} shows that this treatment-dependent ordering matches the Nash equilibria of the reduced game: Always Unsafe and Conditionally Antisocial Safe are behaviourally equivalent equilibrium strategies at low and intermediate $p_r^{\max}$, whereas Conditionally Safe becomes the unique equilibrium at $p_r^{\max}=0.9$.

The model provides a compact explanation for the empirical results. Unsafe technological development is not driven only by fixed individual risk attitudes or by the exogenous risk treatment alone. It can emerge from conditional response dynamics, early aggressive choices, and the competitive pressure created by the race.

\section{Discussion}\label{sec:discussion}

Our two pre-registered hypotheses were not supported: Unsafe behaviour did not differ significantly between the $p_r^{\max}=0.6$ and $0.9$ treatments, and elicited risk preferences did not predict Unsafe choices. The central empirical contribution of this paper instead comes from a set of exploratory analyses that show that Unsafe technological development in this idealised AI race is shaped by interaction history and competitive position. Participants were not simply more or less risk-seeking depending on the exogenous risk treatment or their elicited risk preferences. Instead, they responded to what their opponent had done, to whether they were ahead or behind in the race, and to the early behavioural state of the interaction. This provides behavioural evidence for a central concern in AI and technological race models: competitive pressure can make risky development attractive even when risk is individually costly.

The strongest behavioural predictor of Unsafe play was the opponent's previous action. This suggests that Unsafe development can spread through reciprocal or defensive responses. Once one competitor acts unsafely, the other becomes more likely to do the same, not necessarily because they prefer risk, but because Safe behaviour becomes strategically costly when the competitor is gaining speed. This mechanism is especially relevant in technological races, where safety decisions are rarely made in isolation and where one actor's unsafe behaviour can change the incentives faced by others. It also highlights an important target for public policy in reducing the strategic pressure to match unsafe behaviour by improving transparency, coordination, monitoring, or credible commitments to safe development. This pattern of conditioning on the opponent's previous action is consistent with the reciprocal strategies -- Tit-for-Tat-like reactive rules in particular -- that dominate the strategy classifications inferred in repeated Prisoner's Dilemma experiments \citep{dal2018determinants,monteroPorras2022inferringStrategies}. Our Conditionally Safe and Conditionally Antisocial Safe strategies (\Cref{si:sec:pd-relationship}) can therefore be understood as race-specific instances of this broader family of conditionally cooperative strategies.

Race position further shaped behaviour. Participants who were ahead were less likely to choose Unsafe, whereas those who were behind had stronger incentives to take risks. This supports the idea that fear of falling behind can promote unsafe development. Once more, this effect does not require participants to be generally risk-seeking. Rather, the race structure itself makes Unsafe development attractive when it offers a chance to recover from a disadvantage.

At the same time, participants still reduced Unsafe behaviour when maximum private risk was high. The marginal distributions show a lower modal frequency of Unsafe behaviour for $p_r^{\max}=0.9$, and the evolutionary model predicts a shift towards Conditionally Safe behaviour at high private risk. Thus, high risk does not eliminate Unsafe development, but it changes the strategic conditions under which Unsafe behaviour is favoured. Participants remain willing to take risks when the competitive context makes Unsafe development advantageous, but high private risk weakens the dominance of unconditional Unsafe play.

The reduced evolutionary model provides a mechanistic interpretation of these results. Conditional strategies that respond to the opponent's previous action, especially those that differ in their first-round behaviour, can reproduce the observed treatment-level patterns. At low private risk, Always Unsafe behaviour is favoured because the speed advantage dominates the expected cost of risk. At intermediate risk, a strategy that begins aggressively and then conditions on the opponent's behaviour becomes more successful. At high private risk, Conditionally Safe behaviour becomes dominant, consistent with the observed reduction in Unsafe behaviour. This suggests that Unsafe technological development may be maintained by strategic feedback between early aggressive actions, opponent responses, and race position.

Overall, our results suggest that AI race risks are not only a matter of individual risk preferences. In our data, elicited risk preferences do not significantly predict later Unsafe choices once interaction history and race position are taken into account. This null result mirrors a broader pattern in the repeated Prisoner's Dilemma literature, where elicited risk preferences are typically weak predictors of cooperative strategy choice despite the presence of strategic uncertainty \citep{dal2018determinants}. Instead, Unsafe development arises from competitive dynamics, early behavioural momentum, and strategic responses to rivals. Even when participants face substantial private risk, Unsafe development can persist because the race structure makes safety conditional on what others do. This supports the broader theoretical concern that technological races can produce unsafe outcomes not only because actors underestimate risk, but because competition changes the strategic value of taking that risk.

\section{Limitations of the study}\label{sec:limitations}

The study has several limitations. First, the race is deliberately stylised and short relative to real technological competition. Real AI and technological races involve many actors, repeated public signals, uncertain capabilities, regulatory constraints, and reputational concerns. Second, the reduced model approximates richer distance-dependent behaviour using a small number of deterministic strategies. This makes the mechanism interpretable, but it abstracts away from gradual responses to race distance and from more heterogeneous learning rules. Third, the experimental design focuses on private risk rather than collective or systemic risk. This is useful for isolating individual incentives, but many real safety failures impose externalities on other actors or on society more broadly. Fourth, the central empirical findings reported here -- the roles of opponent behaviour, race position, and first-round momentum -- emerged from exploratory rather than pre-registered analyses. Fifth, the three treatments manipulate only the maximum private risk while holding the competitive structure of the race -- the presence of a rival and the incentive to outpace them -- constant. Sixth, because \(a_i^t\) depends on the entire preceding history of both players' actions rather than solely on \(a_i^{t-1}\) and \(a_{-i}^{t-1}\), the lagged predictors in \Cref{tab:cluster_logit_unsafe} are not strictly exogenous, and the reported associations should not be read as isolating the causal effect of one round's action on the next.

Future work should therefore extend the design to longer horizons, richer strategy spaces, and institutional interventions that can reduce incentives for unsafe development. Particularly important extensions include mechanisms for transparency, commitment, monitoring, liability, or coordinated slowdown. These interventions could test whether reciprocal Unsafe behaviour can be interrupted before it becomes self-reinforcing, and whether lagging competitors can be protected from incentives to catch up through risky development.

\section{Conclusion}\label{sec:conclusion}

This study provides behavioural evidence that Unsafe technological development in an idealised AI race is shaped by competitive dynamics rather than by risk preferences alone. Participants conditioned their choices on the opponent's previous behaviour and on their relative position in the race, with lagging participants showing a stronger tendency to choose Unsafe. A reduced evolutionary model showed that these patterns can be reproduced by a small set of conditional strategies, linking the experimental results to broader theoretical concerns about technological races.

These findings suggest that safety failures in competitive AI development may arise not only because actors underestimate risk, but because competition changes the strategic value of taking that risk. When Unsafe development offers speed and catch-up advantages, safety becomes conditional on the behaviour of rivals. Effective governance of technological races may therefore require interventions that reduce incentives for reciprocal Unsafe behaviour and make Safe development strategically viable even under competitive pressure.

\section{Methods}\label{sec:methods}

\subsection{Experimental design and framing}

The experiment was conducted online using the oTree platform~\citep{chen2016otree}. Participants were recruited from the Prolific~\citep{palan2018prolific} platform. The study was approved by the ``SCEDT Research Ethics sub-Committee'' at School of Computing, Engineering and Digital Technologies, Teesside University, with Review Reference 2023 Oct 16933 Han. All participants provided informed consent.

Participants were assigned to a framed two-player technological race. Each participant took the role of a company developing a technology while competing against another company (see \Cref{si:sec:instructions} for the exact wording of the experimental instructions). In each round, both participants simultaneously chose between Safe and Unsafe development. Participants observed their own and their opponent's previous action, accumulated payoff, accumulated private risk, and current race progress (see \Cref{fig:si:instructions_task2_decision,fig:si:instructions_task2_info}).

The experiment manipulated the maximum private setback risk,
\[
p_r^{\max}\in\{0.1,0.6,0.9\}.
\]
The main behavioural outcome is whether a participant chose Unsafe in a given round. Before the race task, participants completed a risk-preference elicitation task based on the Eckel--Grossman method \citep{eckel2008forecastingRiskAttitudes,dave2010elicitingRiskPreferences}.

The \(p_r^{\max}=0.6\) and \(p_r^{\max}=0.9\) treatments correspond to the preregistered low- and high-risk comparison. The \(p_r^{\max}=0.1\) treatment was added subsequently to test the model prediction that stronger treatment differences should emerge at lower maximum private risk.

\subsection{Repeated AI-race game}

The experimental task is a repeated two-player race. In each round \(t\), player \(i\) chooses
\[
a_i^t\in\{S,U\},
\]
where \(S\) denotes Safe development and \(U\) denotes Unsafe development. Safe development advances the race by one step, whereas Unsafe development advances it by \(1.5\) steps:
\[
\sigma(S)=1,\qquad \sigma(U)=1.5.
\]

The round payoff matrix is
\[
\pi =
\begin{pmatrix}
1 & 0.6\\
2.4 & 2
\end{pmatrix},
\]
where rows denote the focal player's action and columns denote the opponent's action. Thus, Unsafe development provides both a higher immediate payoff and faster race progress.

The game lasts for at least five rounds. After round \(5\), each additional round ends the game with probability \(0.2\). The total number of rounds is therefore
\[
T = 5 + G,
\qquad
G\sim \mathrm{Geom}(0.2)-1,
\]
so that \(\mathbb{E}[T]=9\).

At the end of the race, the player with greater cumulative progress receives a prize of \(100\) ECUs. If both players have equal progress, the prize is split equally. A player's private setback risk depends on the fraction of rounds in which they chose Unsafe:
\[
q_i(T)=p_r^{\max}\frac{n_i^U(T)}{T},
\]
where \(n_i^U(T)\) is the number of Unsafe choices made by player \(i\). Private setback risk applies to players who win the race or tie for first place. If a setback occurs, the affected player loses the payoff from the task. The full specification of this quantity (denoted \(p_r(W)\)), including the tie-handling and no-prize cases, is given in \Cref{si:expected-payoff-matrix}.

In the underlying stage game, Unsafe strictly dominates Safe in every round, as Defect dominates Cooperate in a Prisoner's Dilemma, but mutual Unsafe play is not itself costly at the stage-game level; \Cref{si:sec:pd-relationship} formalises this relationship to the repeated Prisoner's Dilemma and derives the Nash equilibria of the reduced strategy game for each treatment.

\subsection{Statistical analysis}

We model Unsafe choices using a panel logistic regression. The dependent variable equals one if participant \(i\) chose Unsafe in round \(t\). The main predictors are maximum private risk treatment, the participant's previous action \(a_i^{t-1}\), the opponent's previous action \(a_{-i}^{t-1}\), and the race-step difference \(\Delta S_{t-1}\). We also include interaction terms among lagged actions and race-step difference to test whether the effect of race position depends on the previous action profile.

Because the observations are repeated decisions nested within pairs, we report standard errors clustered at the pair level. Models are estimated using observations from round \(2\) onward so that lagged behavioural predictors are defined. Additional specifications include \(a_i^1\), an indicator for whether the focal participant chose Unsafe in the first round, to test whether initial Unsafe behaviour predicts later choices.

\Cref{tab:cluster_logit_unsafe} reports six nested specifications. Models (1)--(3) exclude \(a_i^1\): model (1) includes only the treatment dummies together with the demographic and risk-preference controls; model (2) adds the three lagged predictors (\(a_i^{t-1}\), \(a_{-i}^{t-1}\), \(\Delta S_{t-1}\)) additively; model (3) replaces the additive terms with their full three-way interaction. Models (4)--(6) repeat this same progression after adding \(a_i^1\).

All models include demographic controls for participant sex, age, and nationality, together with risk preferences. Age is centred on its sample mean; nationality is coded as South Africa or Poland, the two dominant recruitment pools (\(44\%\) and \(11\%\) of participants, respectively), relative to all other nationalities, since the full set of \(\sim\!46\) distinct nationalities is too sparse (many singletons) to be estimated as separate dummies. Risk preferences are measured using the Eckel--Grossman elicitation task. The reference treatment is \(p_r^{\max}=0.1\). As is typical for binary-choice models of noisy individual behaviour, McFadden's pseudo \(R^2\) for these models is small in absolute terms (\Cref{tab:cluster_logit_unsafe}); \citet{mcfadden1974conditional} suggests that values in the \(0.2\)--\(0.4\) range already indicate an excellent fit on this scale, so pseudo \(R^2\) here should not be compared to the \(R^2\) of a linear model. The relevant test of the hypotheses is the significance and sign of individual coefficients, not the overall variance explained: substantial round-to-round variation in Unsafe choices is not captured by these covariates, which is expected given the idiosyncratic, individual-level noise inherent in repeated binary decisions.

Because a participant's current action is itself shaped by the full preceding history of both players' choices, not only by the immediately preceding round, \(a_i^{t-1}\) and \(a_{-i}^{t-1}\) are not exogenous regressors. The coefficients reported in \Cref{tab:cluster_logit_unsafe} should therefore be read as conditional associations rather than as estimates of the causal effect of one round's action on the next (\Cref{sec:limitations}).

\subsection{Reduced evolutionary model}

We analyse a reduced strategy set motivated by the regression results:
\begin{itemize}
    \item \textbf{AS} Always Safe: always plays \(S\);
    \item \textbf{AU} Always Unsafe: always plays \(U\);
    \item \textbf{CS} Conditionally Safe: plays \(S\) in round \(1\), then from round 2 copies the opponent's previous action;
    \item \textbf{CAS} Conditionally Antisocial Safe: plays \(U\) in round \(1\), then from round 2 copies the opponent's previous action.
\end{itemize}

For each ordered pair of strategies, we compute the expected total payoff over the stochastic race horizon. Matchups between the two unconditional strategies (AS and AU) admit a closed-form expected payoff obtained by substituting the expected number of rounds. For every matchup involving a conditional strategy (CS or CAS), the expected payoff is instead estimated by Monte Carlo simulation of the stochastic race ($10^4$ replications per matchup), since realised play in these cases depends on the actual sequence of actions taken. The resulting payoff matrix defines the evolutionary game analysed in Figure~\ref{fig:model_predictions}. Payoff derivations and the simulation procedure are described in \Cref{si:expected-payoff-matrix}.

\subsection{Evolutionary dynamics}

We consider a finite well-mixed population of size \(Z\). Individuals revise strategies according to a pairwise comparison process with Fermi imitation. If the expected payoffs of strategies \(A\) and \(B\) are \(f_A\) and \(f_B\), respectively, the probability that an individual using \(B\) adopts \(A\) is
\[
\left(1+\exp[-\beta(f_A-f_B)]\right)^{-1},
\]
where \(\beta\) is the intensity of selection.

Mutation occurs with probability \(\mu\), allowing exploration of alternative strategies. We compute the stationary distribution of the resulting finite-population Markov chain and use it to obtain the predicted average frequency of Unsafe actions. These predictions are compared with the experimental treatment-level frequencies in Figure~\ref{fig:model_predictions}. See \cite{domingos2023egttools} for a more detailed description of the evolutionary model and its implementation in EGTtools.

\section*{Data availability}

All data will be publicly deposited in the same OSF repository of the pre-registration. A private link will be made available for reviewers through the private review link provided in the submission system.

The deposited dataset is a de-identified version of the long-format experimental panel (one row per participant-round), restricted to the fields used in this paper's analyses. It excludes the Prolific participant identifier, all other raw platform and administrative fields, and the free-text post-experiment survey responses. Because the analysed sample ($n=338$; \Cref{si:sec:sample}) is concentrated in a small number of countries (South Africa and Poland account for \(44\%\) and \(11\%\) of participants, respectively), nationality is collapsed to South Africa/Poland/Other -- the same categorisation used in every regression reported here -- and age is reported in five-year bins rather than exactly, removing both as quasi-identifiers without affecting any reported result. Participant and pair identifiers in the deposited data are freshly assigned and do not correspond to any internal study code.

\section*{Code availability}

All analysis code will be deposited in Zenodo and made publicly available upon publication. A private link will be made available for reviewers. The evolutionary simulations were implemented using EGTtools~\citep{domingos2023egttools}. The experimental platform was implemented in oTree~\citep{chen2016otree}.

\section*{Acknowledgements}
E.F.D. is supported by an F.W.O. Senior postdoctoral grant (12A7825N). T.A.H. received funding from the Future of Life Institute (``An empirical study of safe development behaviours in a rapid AI development competition"), and is supported by EPSRC (grant EP/Y00857X/1). The authors thank the helpful comments of Dr. Rémi Suchon which greatly improved the manuscript.

\section*{Author contributions}

E.F.D. and T.A.H. conceived the study. E.F.D. designed and conducted the experiments, developed the online platform, analysed the data, constructed the evolutionary model, and drafted the manuscript. T.A.H. reviewed and revised the data analyses, evolutionary model, and manuscript. All authors discussed the results, contributed to the interpretation of the findings, and approved the final version of the manuscript.

\section*{Competing interests}

The authors declare no competing interests.

\section*{Ethics declarations}

The study was approved by the ``SCEDT Research Ethics sub-Committee'' at School of Computing, Engineering and Digital Technologies, Teesside University with the Review Reference 2023 Oct 16933 Han. All participants provided informed consent before taking part in the experiment.

\bibliography{paper}

%% BioMed_Central_Bib_Style_v1.01

\begin{thebibliography}{40}
% BibTex style file: bmc-mathphys.bst (version 2.1), 2014-07-24
\ifx \bisbn   \undefined \def \bisbn  #1{ISBN #1}\fi
\ifx \binits  \undefined \def \binits#1{#1}\fi
\ifx \bauthor  \undefined \def \bauthor#1{#1}\fi
\ifx \batitle  \undefined \def \batitle#1{#1}\fi
\ifx \bjtitle  \undefined \def \bjtitle#1{#1}\fi
\ifx \bvolume  \undefined \def \bvolume#1{\textbf{#1}}\fi
\ifx \byear  \undefined \def \byear#1{#1}\fi
\ifx \bissue  \undefined \def \bissue#1{#1}\fi
\ifx \bfpage  \undefined \def \bfpage#1{#1}\fi
\ifx \blpage  \undefined \def \blpage #1{#1}\fi
\ifx \burl  \undefined \def \burl#1{\textsf{#1}}\fi
\ifx \doiurl  \undefined \def \doiurl#1{\url{https://doi.org/#1}}\fi
\ifx \betal  \undefined \def \betal{\textit{et al.}}\fi
\ifx \binstitute  \undefined \def \binstitute#1{#1}\fi
\ifx \binstitutionaled  \undefined \def \binstitutionaled#1{#1}\fi
\ifx \bctitle  \undefined \def \bctitle#1{#1}\fi
\ifx \beditor  \undefined \def \beditor#1{#1}\fi
\ifx \bpublisher  \undefined \def \bpublisher#1{#1}\fi
\ifx \bbtitle  \undefined \def \bbtitle#1{#1}\fi
\ifx \bedition  \undefined \def \bedition#1{#1}\fi
\ifx \bseriesno  \undefined \def \bseriesno#1{#1}\fi
\ifx \blocation  \undefined \def \blocation#1{#1}\fi
\ifx \bsertitle  \undefined \def \bsertitle#1{#1}\fi
\ifx \bsnm \undefined \def \bsnm#1{#1}\fi
\ifx \bsuffix \undefined \def \bsuffix#1{#1}\fi
\ifx \bparticle \undefined \def \bparticle#1{#1}\fi
\ifx \barticle \undefined \def \barticle#1{#1}\fi
\bibcommenthead
\ifx \bconfdate \undefined \def \bconfdate #1{#1}\fi
\ifx \botherref \undefined \def \botherref #1{#1}\fi
\ifx \url \undefined \def \url#1{\textsf{#1}}\fi
\ifx \bchapter \undefined \def \bchapter#1{#1}\fi
\ifx \bbook \undefined \def \bbook#1{#1}\fi
\ifx \bcomment \undefined \def \bcomment#1{#1}\fi
\ifx \oauthor \undefined \def \oauthor#1{#1}\fi
\ifx \citeauthoryear \undefined \def \citeauthoryear#1{#1}\fi
\ifx \endbibitem  \undefined \def \endbibitem {}\fi
\ifx \bconflocation  \undefined \def \bconflocation#1{#1}\fi
\ifx \arxivurl  \undefined \def \arxivurl#1{\textsf{#1}}\fi
\csname PreBibitemsHook\endcsname

%%% 1
\bibitem[\protect\citeauthoryear{Aghion
  et~al.}{2005}]{AghionEtAl2005CompetitionInnovation}
\begin{barticle}
\bauthor{\bsnm{Aghion}, \binits{P.}},
\bauthor{\bsnm{Bloom}, \binits{N.}},
\bauthor{\bsnm{Blundell}, \binits{R.}},
\bauthor{\bsnm{Griffith}, \binits{R.}},
\bauthor{\bsnm{Howitt}, \binits{P.}}:
\batitle{Competition and innovation: An inverted-u relationship}.
\bjtitle{The Quarterly Journal of Economics}
\bvolume{120}(\bissue{2}),
\bfpage{701}--\blpage{728}
(\byear{2005})
\doiurl{10.1162/0033553053970214}
\end{barticle}
\endbibitem

%%% 2
\bibitem[\protect\citeauthoryear{Askell
  et~al.}{2019}]{AskellBrundageHadfield2019CooperationResponsibleAI}
\begin{barticle}
\bauthor{\bsnm{Askell}, \binits{A.}},
\bauthor{\bsnm{Brundage}, \binits{M.}},
\bauthor{\bsnm{Hadfield}, \binits{G.K.}}:
\batitle{The role of cooperation in responsible {AI} development}.
\bjtitle{arXiv preprint arXiv:1907.04534}
(\byear{2019})
\doiurl{10.48550/arXiv.1907.04534}
{\href{https://arxiv.org/abs/1907.04534}{{arXiv:1907.04534}}}
{[cs.CY]}
\end{barticle}
\endbibitem

%%% 3
\bibitem[\protect\citeauthoryear{Armstrong
  et~al.}{2016}]{ArmstrongBostromShulman2016RacingPrecipice}
\begin{barticle}
\bauthor{\bsnm{Armstrong}, \binits{S.}},
\bauthor{\bsnm{Bostrom}, \binits{N.}},
\bauthor{\bsnm{Shulman}, \binits{C.}}:
\batitle{Racing to the precipice: A model of artificial intelligence
  development}.
\bjtitle{AI \& Society}
\bvolume{31}(\bissue{2}),
\bfpage{201}--\blpage{206}
(\byear{2016})
\doiurl{10.1007/s00146-015-0590-y}
\end{barticle}
\endbibitem

%%% 4
\bibitem[\protect\citeauthoryear{{Bueno de Mesquita}
  et~al.}{2026}]{BuenoDeMesquitaDziudaPolborn2026AGIRace}
\begin{botherref}
\oauthor{\bsnm{{Bueno de Mesquita}}, \binits{E.}},
\oauthor{\bsnm{Dziuda}, \binits{W.}},
\oauthor{\bsnm{Polborn}, \binits{M.}}:
The {AGI} race and existential risk.
Technical Report 35276,
National Bureau of Economic Research
(May 2026).
\doiurl{10.3386/w35276}
\end{botherref}
\endbibitem

%%% 5
\bibitem[\protect\citeauthoryear{Bengio et~al.}{2024}]{bengio2024managing}
\begin{barticle}
\bauthor{\bsnm{Bengio}, \binits{Y.}},
\bauthor{\bsnm{Hinton}, \binits{G.}},
\bauthor{\bsnm{Yao}, \binits{A.}},
\bauthor{\bsnm{Song}, \binits{D.}},
\bauthor{\bsnm{Abbeel}, \binits{P.}},
\bauthor{\bsnm{Darrell}, \binits{T.}},
\bauthor{\bsnm{Harari}, \binits{Y.N.}},
\bauthor{\bsnm{Zhang}, \binits{Y.-Q.}},
\bauthor{\bsnm{Xue}, \binits{L.}},
\bauthor{\bsnm{Shalev-Shwartz}, \binits{S.}}, \betal:
\batitle{Managing extreme {AI} risks amid rapid progress}.
\bjtitle{Science}
\bvolume{384}(\bissue{6698}),
\bfpage{842}--\blpage{845}
(\byear{2024})
\end{barticle}
\endbibitem

%%% 6
\bibitem[\protect\citeauthoryear{Brown}{2021}]{brown2021linearMixedEffectsR}
\begin{barticle}
\bauthor{\bsnm{Brown}, \binits{V.A.}}:
\batitle{An introduction to linear mixed-effects modeling in {R}}.
\bjtitle{Advances in Methods and Practices in Psychological Science}
\bvolume{4}(\bissue{1}),
\bfpage{2515245920960351}
(\byear{2021})
\doiurl{10.1177/2515245920960351}
\end{barticle}
\endbibitem

%%% 7
\bibitem[\protect\citeauthoryear{Barnett and
  Scher}{2025}]{BarnettScher2025AIGovernanceExtinction}
\begin{botherref}
\oauthor{\bsnm{Barnett}, \binits{P.}},
\oauthor{\bsnm{Scher}, \binits{A.}}:
{AI} governance to avoid extinction: The strategic landscape and actionable
  research questions.
arXiv preprint arXiv:2505.04592
(2025)
\end{botherref}
\endbibitem

%%% 8
\bibitem[\protect\citeauthoryear{Bull
  et~al.}{1987}]{BullSchotterWeigelt1987TournamentsPieceRates}
\begin{barticle}
\bauthor{\bsnm{Bull}, \binits{C.}},
\bauthor{\bsnm{Schotter}, \binits{A.}},
\bauthor{\bsnm{Weigelt}, \binits{K.}}:
\batitle{Tournaments and piece rates: An experimental study}.
\bjtitle{Journal of Political Economy}
\bvolume{95}(\bissue{1}),
\bfpage{1}--\blpage{33}
(\byear{1987})
\doiurl{10.1086/261439}
\end{barticle}
\endbibitem

%%% 9
\bibitem[\protect\citeauthoryear{Cave and
  {\'{O}}h{\'{E}}igeartaigh}{2018}]{CaveOHeigeartaigh2018AIRace}
\begin{bchapter}
\bauthor{\bsnm{Cave}, \binits{S.}},
\bauthor{\bsnm{{\'{O}}h{\'{E}}igeartaigh}, \binits{S.S.}}:
\bctitle{An {AI} race for strategic advantage: Rhetoric and risks}.
In: \bbtitle{Proceedings of the 2018 AAAI/ACM Conference on AI, Ethics, and
  Society},
pp. \bfpage{36}--\blpage{40}.
\bpublisher{Association for Computing Machinery},
\blocation{New York, NY, USA}
(\byear{2018}).
\doiurl{10.1145/3278721.3278780}
\end{bchapter}
\endbibitem

%%% 10
\bibitem[\protect\citeauthoryear{Cimpeanu
  et~al.}{2022}]{CimpeanuEtAl2022AIRacesHeterogeneous}
\begin{barticle}
\bauthor{\bsnm{Cimpeanu}, \binits{T.}},
\bauthor{\bsnm{Santos}, \binits{F.C.}},
\bauthor{\bsnm{Pereira}, \binits{L.M.}},
\bauthor{\bsnm{Lenaerts}, \binits{T.}},
\bauthor{\bsnm{Han}, \binits{T.A.}}:
\batitle{Artificial intelligence development races in heterogeneous settings}.
\bjtitle{Scientific Reports}
\bvolume{12},
\bfpage{1723}
(\byear{2022})
\doiurl{10.1038/s41598-022-05729-3}
\end{barticle}
\endbibitem

%%% 11
\bibitem[\protect\citeauthoryear{Chen et~al.}{2016}]{chen2016otree}
\begin{barticle}
\bauthor{\bsnm{Chen}, \binits{D.L.}},
\bauthor{\bsnm{Schonger}, \binits{M.}},
\bauthor{\bsnm{Wickens}, \binits{C.}}:
\batitle{otree—an open-source platform for laboratory, online, and field
  experiments}.
\bjtitle{Journal of Behavioral and Experimental Finance}
\bvolume{9},
\bfpage{88}--\blpage{97}
(\byear{2016})
\end{barticle}
\endbibitem

%%% 12
\bibitem[\protect\citeauthoryear{Dal~B{\'o} and
  Fr{\'e}chette}{2018}]{dal2018determinants}
\begin{barticle}
\bauthor{\bsnm{Dal~B{\'o}}, \binits{P.}},
\bauthor{\bsnm{Fr{\'e}chette}, \binits{G.R.}}:
\batitle{On the determinants of cooperation in infinitely repeated games: A
  survey}.
\bjtitle{Journal of Economic Literature}
\bvolume{56}(\bissue{1}),
\bfpage{60}--\blpage{114}
(\byear{2018})
\end{barticle}
\endbibitem

%%% 13
\bibitem[\protect\citeauthoryear{Dave
  et~al.}{2010}]{dave2010elicitingRiskPreferences}
\begin{barticle}
\bauthor{\bsnm{Dave}, \binits{C.}},
\bauthor{\bsnm{Eckel}, \binits{C.C.}},
\bauthor{\bsnm{Johnson}, \binits{C.A.}},
\bauthor{\bsnm{Rojas}, \binits{C.}}:
\batitle{Eliciting risk preferences: When is simple better?}
\bjtitle{Journal of Risk and Uncertainty}
\bvolume{41}(\bissue{3}),
\bfpage{219}--\blpage{243}
(\byear{2010})
\doiurl{10.1007/s11166-010-9103-z}
\end{barticle}
\endbibitem

%%% 14
\bibitem[\protect\citeauthoryear{Domingos et~al.}{2023}]{domingos2023egttools}
\begin{botherref}
\oauthor{\bsnm{Domingos}, \binits{E.F.}},
\oauthor{\bsnm{Santos}, \binits{F.C.}},
\oauthor{\bsnm{Lenaerts}, \binits{T.}}:
Egttools: Evolutionary game dynamics in python.
Iscience
\textbf{26}(4)
(2023)
\end{botherref}
\endbibitem

%%% 15
\bibitem[\protect\citeauthoryear{Eckel and
  Grossman}{2008}]{eckel2008forecastingRiskAttitudes}
\begin{barticle}
\bauthor{\bsnm{Eckel}, \binits{C.C.}},
\bauthor{\bsnm{Grossman}, \binits{P.J.}}:
\batitle{Forecasting risk attitudes: An experimental study using actual and
  forecast gamble choices}.
\bjtitle{Journal of Economic Behavior \& Organization}
\bvolume{68}(\bissue{1}),
\bfpage{1}--\blpage{17}
(\byear{2008})
\doiurl{10.1016/j.jebo.2008.04.006}
\end{barticle}
\endbibitem

%%% 16
\bibitem[\protect\citeauthoryear{Eriksen and
  Kval{\o}y}{2014}]{EriksenKvaloy2014MyopicRiskTaking}
\begin{barticle}
\bauthor{\bsnm{Eriksen}, \binits{K.W.}},
\bauthor{\bsnm{Kval{\o}y}, \binits{O.}}:
\batitle{Myopic risk-taking in tournaments}.
\bjtitle{Journal of Economic Behavior \& Organization}
\bvolume{97},
\bfpage{37}--\blpage{46}
(\byear{2014})
\doiurl{10.1016/j.jebo.2013.10.004}
\end{barticle}
\endbibitem

%%% 17
\bibitem[\protect\citeauthoryear{Emery-Xu
  et~al.}{2024}]{EmeryXuParkTrager2024TechnologyRaces}
\begin{barticle}
\bauthor{\bsnm{Emery-Xu}, \binits{N.}},
\bauthor{\bsnm{Park}, \binits{A.}},
\bauthor{\bsnm{Trager}, \binits{R.}}:
\batitle{Uncertainty, information, and risk in international technology races}.
\bjtitle{Journal of Conflict Resolution}
\bvolume{68}(\bissue{10}),
\bfpage{2019}--\blpage{2047}
(\byear{2024})
\doiurl{10.1177/00220027231214996}
\end{barticle}
\endbibitem

%%% 18
\bibitem[\protect\citeauthoryear{Filippin and
  Gioia}{2018}]{FilippinGioia2018CompetitionSubsequentRiskTaking}
\begin{barticle}
\bauthor{\bsnm{Filippin}, \binits{A.}},
\bauthor{\bsnm{Gioia}, \binits{F.}}:
\batitle{Competition and subsequent risk-taking behaviour: Heterogeneity across
  gender and outcomes}.
\bjtitle{Journal of Behavioral and Experimental Economics}
\bvolume{75},
\bfpage{84}--\blpage{94}
(\byear{2018})
\doiurl{10.1016/j.socec.2018.05.003}
\end{barticle}
\endbibitem

%%% 19
\bibitem[\protect\citeauthoryear{Fudenberg and
  Tirole}{1985}]{FudenbergTirole1985Preemption}
\begin{barticle}
\bauthor{\bsnm{Fudenberg}, \binits{D.}},
\bauthor{\bsnm{Tirole}, \binits{J.}}:
\batitle{Preemption and rent equalization in the adoption of new technology}.
\bjtitle{The Review of Economic Studies}
\bvolume{52}(\bissue{3}),
\bfpage{383}--\blpage{401}
(\byear{1985})
\doiurl{10.2307/2297660}
\end{barticle}
\endbibitem

%%% 20
\bibitem[\protect\citeauthoryear{Gruetzemacher
  et~al.}{2025}]{gruetzemacher2025strategic}
\begin{barticle}
\bauthor{\bsnm{Gruetzemacher}, \binits{R.}},
\bauthor{\bsnm{Avin}, \binits{S.}},
\bauthor{\bsnm{Fox}, \binits{J.}},
\bauthor{\bsnm{Saeri}, \binits{A.K.}}:
\batitle{Strategic insights from simulation gaming of {AI} race dynamics}.
\bjtitle{Futures}
\bvolume{167},
\bfpage{103563}
(\byear{2025})
\end{barticle}
\endbibitem

%%% 21
\bibitem[\protect\citeauthoryear{Genakos and
  Pagliero}{2012}]{GenakosPagliero2012InterimRankRiskTaking}
\begin{barticle}
\bauthor{\bsnm{Genakos}, \binits{C.}},
\bauthor{\bsnm{Pagliero}, \binits{M.}}:
\batitle{Interim rank, risk taking, and performance in dynamic tournaments}.
\bjtitle{Journal of Political Economy}
\bvolume{120}(\bissue{4}),
\bfpage{782}--\blpage{813}
(\byear{2012})
\doiurl{10.1086/668502}
\end{barticle}
\endbibitem

%%% 22
\bibitem[\protect\citeauthoryear{G{\"u}rtler
  et~al.}{2023}]{GuertlerStruthThon2023CompetitionRiskTaking}
\begin{barticle}
\bauthor{\bsnm{G{\"u}rtler}, \binits{O.}},
\bauthor{\bsnm{Struth}, \binits{L.}},
\bauthor{\bsnm{Thon}, \binits{M.}}:
\batitle{Competition and risk-taking}.
\bjtitle{European Economic Review}
\bvolume{160},
\bfpage{104592}
(\byear{2023})
\doiurl{10.1016/j.euroecorev.2023.104592}
\end{barticle}
\endbibitem

%%% 23
\bibitem[\protect\citeauthoryear{Hoenig and Heisey}{2001}]{hoenig2001abuse}
\begin{barticle}
\bauthor{\bsnm{Hoenig}, \binits{J.M.}},
\bauthor{\bsnm{Heisey}, \binits{D.M.}}:
\batitle{The abuse of power: The pervasive fallacy of power calculations for
  data analysis}.
\bjtitle{The American Statistician}
\bvolume{55}(\bissue{1}),
\bfpage{19}--\blpage{24}
(\byear{2001})
\doiurl{10.1198/000313001300339897}
\end{barticle}
\endbibitem

%%% 24
\bibitem[\protect\citeauthoryear{Han et~al.}{2022}]{han2022voluntary}
\begin{barticle}
\bauthor{\bsnm{Han}, \binits{T.A.}},
\bauthor{\bsnm{Lenaerts}, \binits{T.}},
\bauthor{\bsnm{Santos}, \binits{F.C.}},
\bauthor{\bsnm{Pereira}, \binits{L.M.}}:
\batitle{Voluntary safety commitments provide an escape from over-regulation in
  {AI} development}.
\bjtitle{Technology in Society}
\bvolume{68},
\bfpage{101843}
(\byear{2022})
\end{barticle}
\endbibitem

%%% 25
\bibitem[\protect\citeauthoryear{Han et~al.}{2020}]{han2020toRegulateOrNot}
\begin{barticle}
\bauthor{\bsnm{Han}, \binits{T.A.}},
\bauthor{\bsnm{Moniz~Pereira}, \binits{L.}},
\bauthor{\bsnm{Santos}, \binits{F.C.}},
\bauthor{\bsnm{Lenaerts}, \binits{T.}}:
\batitle{To regulate or not: A social dynamics analysis of an idealised {AI}
  race}.
\bjtitle{Journal of Artificial Intelligence Research}
\bvolume{69},
\bfpage{881}--\blpage{921}
(\byear{2020})
\doiurl{10.1613/jair.1.12225}
\end{barticle}
\endbibitem

%%% 26
\bibitem[\protect\citeauthoryear{Hendrycks
  et~al.}{2023}]{hendrycks2023overview}
\begin{botherref}
\oauthor{\bsnm{Hendrycks}, \binits{D.}},
\oauthor{\bsnm{Mazeika}, \binits{M.}},
\oauthor{\bsnm{Woodside}, \binits{T.}}:
An overview of catastrophic {AI} risks.
arXiv preprint arXiv:2306.12001
(2023)
\end{botherref}
\endbibitem

%%% 27
\bibitem[\protect\citeauthoryear{Han et~al.}{2021}]{HanEtAl2021MediatingAI}
\begin{barticle}
\bauthor{\bsnm{Han}, \binits{T.A.}},
\bauthor{\bsnm{Pereira}, \binits{L.M.}},
\bauthor{\bsnm{Lenaerts}, \binits{T.}},
\bauthor{\bsnm{Santos}, \binits{F.C.}}:
\batitle{Mediating artificial intelligence developments through negative and
  positive incentives}.
\bjtitle{PLOS ONE}
\bvolume{16}(\bissue{1}),
\bfpage{0244592}
(\byear{2021})
\doiurl{10.1371/journal.pone.0244592}
\end{barticle}
\endbibitem

%%% 28
\bibitem[\protect\citeauthoryear{Loury}{1979}]{Loury1979MarketStructureInnovation}
\begin{barticle}
\bauthor{\bsnm{Loury}, \binits{G.C.}}:
\batitle{Market structure and innovation}.
\bjtitle{The Quarterly Journal of Economics}
\bvolume{93}(\bissue{3}),
\bfpage{395}--\blpage{410}
(\byear{1979})
\end{barticle}
\endbibitem

%%% 29
\bibitem[\protect\citeauthoryear{McFadden}{1974}]{mcfadden1974conditional}
\begin{bchapter}
\bauthor{\bsnm{McFadden}, \binits{D.}}:
\bctitle{Conditional logit analysis of qualitative choice behavior}.
In: \beditor{\bsnm{Zarembka}, \binits{P.}} (ed.)
\bbtitle{Frontiers in Econometrics},
pp. \bfpage{105}--\blpage{142}.
\bpublisher{Academic Press},
\blocation{New York}
(\byear{1974})
\end{bchapter}
\endbibitem

%%% 30
\bibitem[\protect\citeauthoryear{Montero-Porras
  et~al.}{2022}]{monteroPorras2022inferringStrategies}
\begin{barticle}
\bauthor{\bsnm{Montero-Porras}, \binits{E.}},
\bauthor{\bsnm{Gruji{\'c}}, \binits{J.}},
\bauthor{\bsnm{Fern{\'a}ndez~Domingos}, \binits{E.}},
\bauthor{\bsnm{Lenaerts}, \binits{T.}}:
\batitle{Inferring strategies from observations in long iterated prisoner's
  dilemma experiments}.
\bjtitle{Scientific Reports}
\bvolume{12}(\bissue{1}),
\bfpage{7589}
(\byear{2022})
\doiurl{10.1038/s41598-022-11654-2}
\end{barticle}
\endbibitem

%%% 31
\bibitem[\protect\citeauthoryear{M{\"u}ller et~al.}{2026}]{muller2026evolvable}
\begin{barticle}
\bauthor{\bsnm{M{\"u}ller}, \binits{V.}},
\bauthor{\bsnm{Steels}, \binits{L.}},
\bauthor{\bsnm{Szathm{\'a}ry}, \binits{E.}}:
\batitle{Evolvable {AI}: Threats of a new major transition in evolution}.
\bjtitle{Proceedings of the National Academy of Sciences}
\bvolume{123}(\bissue{17}),
\bfpage{2527700123}
(\byear{2026})
\doiurl{10.1073/pnas.2527700123}
\end{barticle}
\endbibitem

%%% 32
\bibitem[\protect\citeauthoryear{Nieken and
  Sliwka}{2010}]{NiekenSliwka2010RiskTakingTournaments}
\begin{barticle}
\bauthor{\bsnm{Nieken}, \binits{P.}},
\bauthor{\bsnm{Sliwka}, \binits{D.}}:
\batitle{Risk-taking tournaments: Theory and experimental evidence}.
\bjtitle{Journal of Economic Psychology}
\bvolume{31}(\bissue{3}),
\bfpage{254}--\blpage{268}
(\byear{2010})
\doiurl{10.1016/j.joep.2009.03.009}
\end{barticle}
\endbibitem

%%% 33
\bibitem[\protect\citeauthoryear{Ord}{2020}]{ord2020precipice}
\begin{bbook}
\bauthor{\bsnm{Ord}, \binits{T.}}:
\bbtitle{The Precipice: Existential Risk and the Future of Humanity}.
\bpublisher{Hachette UK},
\blocation{London}
(\byear{2020})
\end{bbook}
\endbibitem

%%% 34
\bibitem[\protect\citeauthoryear{Polborn}{2025}]{Polborn2025CompetitiveDevelopmentDangerousTechnologies}
\begin{botherref}
\oauthor{\bsnm{Polborn}, \binits{M.K.}}:
Competitive development of dangerous technologies.
Technical report,
Vanderbilt University
(April 2025)
\end{botherref}
\endbibitem

%%% 35
\bibitem[\protect\citeauthoryear{Palan and Schitter}{2018}]{palan2018prolific}
\begin{barticle}
\bauthor{\bsnm{Palan}, \binits{S.}},
\bauthor{\bsnm{Schitter}, \binits{C.}}:
\batitle{Prolific.ac—a subject pool for online experiments}.
\bjtitle{Journal of Behavioral and Experimental Finance}
\bvolume{17},
\bfpage{22}--\blpage{27}
(\byear{2018})
\doiurl{10.1016/j.jbef.2017.12.004}
\end{barticle}
\endbibitem

%%% 36
\bibitem[\protect\citeauthoryear{Saeri et~al.}{2026}]{saeri2026prioritization}
\begin{botherref}
\oauthor{\bsnm{Saeri}, \binits{A.K.}},
\oauthor{\bsnm{Graham}, \binits{J.}},
\oauthor{\bsnm{Noetel}, \binits{M.}},
\oauthor{\bsnm{Slattery}, \binits{P.}},
\oauthor{\bsnm{Ah-King}, \binits{D.}},
\oauthor{\bsnm{Aittokallio}, \binits{E.}},
\oauthor{\bsnm{Akindehin}, \binits{I.}},
\oauthor{\bsnm{Mahdi}, \binits{A.A.}},
\oauthor{\bsnm{Alhajjar}, \binits{E.}},
\oauthor{\bsnm{Lipcsey}, \binits{R.A.}}, et al.:
Prioritization of risks from artificial intelligence: A delphi study of 272
  international experts.
arXiv preprint arXiv:2606.04490
(2026)
\end{botherref}
\endbibitem

%%% 37
\bibitem[\protect\citeauthoryear{Schedlinsky
  et~al.}{2016}]{SchedlinskySommerWoehrmann2016RiskTakingTournaments}
\begin{barticle}
\bauthor{\bsnm{Schedlinsky}, \binits{I.}},
\bauthor{\bsnm{Sommer}, \binits{F.}},
\bauthor{\bsnm{W{\"o}hrmann}, \binits{A.}}:
\batitle{Risk-taking in tournaments: an experimental analysis}.
\bjtitle{Journal of Business Economics}
\bvolume{86}(\bissue{8}),
\bfpage{837}--\blpage{866}
(\byear{2016})
\doiurl{10.1007/s11573-016-0813-9}
\end{barticle}
\endbibitem

%%% 38
\bibitem[\protect\citeauthoryear{Stafford
  et~al.}{2022}]{StaffordTragerDafoe2022SafetyNotGuaranteed}
\begin{botherref}
\oauthor{\bsnm{Stafford}, \binits{E.}},
\oauthor{\bsnm{Trager}, \binits{R.F.}},
\oauthor{\bsnm{Dafoe}, \binits{A.}}:
Safety not guaranteed: International races for risky technologies.
Technical report,
Centre for the Governance of AI
(November 2022)
\end{botherref}
\endbibitem

%%% 39
\bibitem[\protect\citeauthoryear{Spadoni
  et~al.}{2018}]{Spadoni2018CompetitionRiskTakingContests}
\begin{barticle}
\bauthor{\bsnm{Spadoni}, \binits{L.}},
\bauthor{\bsnm{Ven}, \binits{J.}},
\bauthor{\bsnm{Willems}, \binits{T.}}:
\batitle{The effect of competition on risk taking in contests}.
\bjtitle{Games}
\bvolume{9}(\bissue{3}),
\bfpage{72}
(\byear{2018})
\doiurl{10.3390/g9030072}
\end{barticle}
\endbibitem

%%% 40
\bibitem[\protect\citeauthoryear{Zizzo}{2002}]{Zizzo2002RacingWithUncertainty}
\begin{barticle}
\bauthor{\bsnm{Zizzo}, \binits{D.J.}}:
\batitle{Racing with uncertainty: a patent race experiment}.
\bjtitle{International Journal of Industrial Organization}
\bvolume{20},
\bfpage{877}--\blpage{902}
(\byear{2002})
\doiurl{10.1016/S0167-7187(01)00087-X}
\end{barticle}
\endbibitem

\end{thebibliography}

\clearpage
\section*{Supporting Information}
\addcontentsline{toc}{section}{Supporting Information}

\renewcommand{\thesection}{S\arabic{section}}
\renewcommand{\thesubsection}{S\arabic{section}.\arabic{subsection}}
\renewcommand{\thesubsubsection}{S\arabic{section}.\arabic{subsection}.\arabic{subsubsection}}
\renewcommand{\thefigure}{S\arabic{figure}}
\renewcommand{\thetable}{S\arabic{table}}

\makeatletter
\renewcommand{\theHsection}{SI.\arabic{section}}
\renewcommand{\theHsubsection}{SI.\arabic{section}.\arabic{subsection}}
\renewcommand{\theHsubsubsection}{SI.\arabic{section}.\arabic{subsection}.\arabic{subsubsection}}
\renewcommand{\theHfigure}{SI.\arabic{figure}}
\renewcommand{\theHtable}{SI.\arabic{table}}
\makeatother

\setcounter{section}{0}
\setcounter{subsection}{0}
\setcounter{subsubsection}{0}
\setcounter{figure}{0}
\setcounter{table}{0}

\section{Supplementary experimental material}
\label{si:sec:experimental-material}

\subsection{Pre-registration}
\label{si:sec:prereg}

This study was pre-registered on OSF on 24/05/2024. The pre-registration document is available at \href{https://osf.io/pzyfm}{https://osf.io/pzyfm}.

\subsubsection{Pre-registered hypotheses}
\label{si:sec:prereg-hypotheses}

\paragraph{Q1: How does the maximum level of private risk affect the frequency of Safe decisions for technology development?}

\begin{enumerate}[label={Hypothesis 1.\arabic*:}, labelwidth={90pt}, leftmargin=*]
    \item The maximum level of private risk affects the odds of acting Safe.
    \begin{description}
        \item[\textbf{\(H_0\) 1.1:}] The maximum level of private risk does not affect the odds of acting Safe.
        \item[\textbf{\(H_A\) 1.1:}] The maximum level of private risk affects the odds of acting Safe.
    \end{description}

    \item High maximum private risk (\(p_r^{\max}=0.9\)) increases the odds of acting Safe with respect to low maximum private risk (\(p_r^{\max}=0.6\)).
    \begin{description}
        \item[\textbf{\(H_0\) 1.2:}] High maximum private risk does not increase the odds of acting Safe with respect to low maximum private risk.
        \item[\textbf{\(H_A\) 1.2:}] High maximum private risk increases the odds of acting Safe with respect to low maximum private risk.
    \end{description}
\end{enumerate}

\paragraph{Q2: Are risk preferences correlated with the frequency of Unsafe choices?}

Following the original pre-registration, this question was framed around the two originally pre-registered treatments, referred to below as T1 (\(p_r^{\max}=0.9\), the high-risk treatment) and T2 (\(p_r^{\max}=0.6\), the low-risk treatment); the subsequently added \(p_r^{\max}=0.1\) treatment (\Cref{si:sec:prereg-changes}) was not part of this comparison.

\begin{enumerate}[label={Hypothesis 2.\arabic*:}, labelwidth={90pt}, leftmargin=*]
    \item Participants' risk preferences influence the odds of acting Unsafe in T1.
    \begin{description}
        \item[\textbf{\(H_0\) 2.1:}] Participants' risk preferences do not influence the odds of acting Unsafe in T1.
        \item[\textbf{\(H_A\) 2.1:}] Participants' risk preferences influence the odds of acting Unsafe in T1.
    \end{description}

    \item Participants' risk preferences influence the odds of acting Unsafe in T2.
    \begin{description}
        \item[\textbf{\(H_0\) 2.2:}] Participants' risk preferences do not influence the odds of acting Unsafe in T2.
        \item[\textbf{\(H_A\) 2.2:}] Participants' risk preferences influence the odds of acting Unsafe in T2.
    \end{description}

    \item Risk preferences have a stronger influence on the odds of acting Safe in the high-risk treatment T1 than in the low-risk treatment T2.
    \begin{description}
        \item[\textbf{\(H_0\) 2.3:}] The influence of risk preferences on the odds of acting Safe is not stronger in T1 than in T2.
        \item[\textbf{\(H_A\) 2.3:}] The influence of risk preferences on the odds of acting Safe is stronger in T1 than in T2.
    \end{description}
\end{enumerate}

\paragraph{Q3: Does an increase in private risk affect the distribution of strategies in the experimental population?}

This was an exploratory question. Addressing it requires inferring the distribution of strategies used by participants in the experiment and determining how closely these inferred strategies correspond to the strategies proposed by Han et al.~\citep{han2020toRegulateOrNot}. We use this analysis to inform the design of future follow-up empirical studies.

\subsubsection{Pre-registered analyses}
\label{si:sec:prereg-analyses}

In the main text, we report fixed-effects cluster-robust panel logistic regressions (\Cref{tab:cluster_logit_unsafe}). Here we report the pre-registered mixed-effects regression \citep{brown2021linearMixedEffectsR} testing whether risk preferences predict the frequency of Unsafe decisions across treatments (Hypotheses 2.1--2.3). Results are reported in \Cref{tab:si:prereg-mixed-model}. As in \Cref{tab:cluster_logit_unsafe}, the mixed-effects model shows that the most robust predictor of Unsafe choices is the participant's own previous action, followed by the opponent's previous action. Risk preference does not significantly predict Unsafe choices, either as a main effect or through its interaction with maximum private risk, consistent with Hypotheses 2.1--2.3 not being supported by the data.

\begin{table}[h]
\centering
\caption{\textbf{Mixed-effects logistic regression predicting Unsafe choices.} The model predicts the probability of choosing Unsafe with a random intercept for group identity. Maximum private risk is entered as a continuous score, centred on its sample mean. Risk preference is the participant's Eckel--Grossman gamble choice (0--5), entered as a continuous score since the six gambles have monotonically increasing payoff risk by design.}
\label{tab:si:prereg-mixed-model}
\begin{tabular}{lrrrr}
\toprule
\textbf{Fixed effect} & \textbf{Estimate} & \textbf{Std. Error} & \textbf{z value} & \textbf{Pr(\(>|z|\))} \\
\midrule
Intercept & 0.5949 & 0.1425 & 4.175 & \(<0.001^{***}\) \\
Max.\ private risk & -0.1234 & 0.3098 & -0.398 & \(0.6903\) \\
\(a_i(t-1)\): Unsafe & -0.6203 & 0.1471 & -4.216 & \(<0.001^{***}\) \\
\(a_{-i}(t-1)\): Unsafe & 0.2820 & 0.1480 & 1.905 & \(0.0567^{\dagger}\) \\
\(\Delta S(t-1)\) & -0.2863 & 0.1546 & -1.851 & \(0.0641^{\dagger}\) \\
Risk preference & -0.0452 & 0.0312 & -1.447 & \(0.1480\) \\
Unsafe \(\times\) Unsafe & 0.3046 & 0.1986 & 1.534 & \(0.1250\) \\
Unsafe \(\times\) \(\Delta S(t-1)\) & 0.3496 & 0.2019 & 1.732 & \(0.0833^{\dagger}\) \\
Opponent Unsafe \(\times\) \(\Delta S(t-1)\) & 0.3923 & 0.2077 & 1.889 & \(0.0589^{\dagger}\) \\
Three-way interaction & -0.1588 & 0.2582 & -0.615 & \(0.5384\) \\
Risk preference \(\times\) Max.\ private risk & -0.1354 & 0.0981 & -1.380 & \(0.1675\) \\
\bottomrule
\end{tabular}

\vspace{0.5em}
\raggedright
\footnotesize Note. Random intercept for group identity: SD = 0.8088. Significance codes: \(^{\dagger}p<0.1\), \(^{*}p<0.05\), \(^{**}p<0.01\), \(^{***}p<0.001\).
\end{table}

\subsubsection{Changes to the pre-registration}
\label{si:sec:prereg-changes}

Although our experimental hypotheses were stated in terms of Safe decisions, the action space is binary. Therefore, the frequency of Safe decisions satisfies \(\phi_S=1-\phi_U\), and we report the main results in terms of the frequency of Unsafe decisions \(\phi_U\).

We found no significant effect of elicited risk preferences in the main experiment. Therefore, we mention this result in the main text and report the corresponding robustness analyses in \Cref{si:sec:prereg-analyses} and \Cref{si:sec:data-analysis-risk}.

The initial pre-registration included treatments \(p_r^{\max}\in\{0.6,0.9\}\), collected in the first data-collection wave (\Cref{tab:si:collection-dates}). After observing that participants' behaviour differed from the original model predictions, in particular because many participants chose Unsafe in the first round while later behaving conditionally, we added a lower-risk treatment \(p_r^{\max}=0.1\), collected in a second wave beginning 2024-11-11. This additional treatment was motivated by the updated model, which predicts clearer treatment differences at lower maximum private risk.

\paragraph{Observed effect sizes for the pre-registered comparison.} The pre-registered comparison (Hypothesis 1.2) is between the \(p_r^{\max}=0.6\) and \(p_r^{\max}=0.9\) treatments. Using the same round-\(t\geq2\), covariate-complete analysed sample as \Cref{tab:cluster_logit_unsafe} (realised participant-level mean Unsafe frequencies over round \(t\geq2\); \(N=105\) and \(N=136\), respectively; \Cref{tab:si:summary-stats}), the observed effect size for this comparison is negligible (Cohen's \(d=-0.027\), independent-samples \(t=-0.206\), \(p=0.837\)), indicating that the true effect of the originally hypothesised \(0.6\)-vs-\(0.9\) contrast, if any, is small. By contrast, the comparisons between the added \(p_r^{\max}=0.1\) treatment and each of the pre-registered treatments show a small-to-moderate observed effect (\(p_r^{\max}=0.1\) vs.\ \(0.6\): \(d=0.341\); \(p_r^{\max}=0.1\) vs.\ \(0.9\): \(d=0.323\)). These effect sizes are computed on the round-\(t\geq2\), covariate-complete analysed sample and therefore differ slightly from the raw, all-rounds per-participant comparison underlying the significance tests reported in \Cref{fig:treatment_differences}A, which additionally include round-\(1\) decisions and participants without complete demographic or risk-preference covariates. We do not report post-hoc statistical power for these comparisons, since power computed from an observed effect size is a deterministic function of the corresponding \(p\)-value and does not provide independent evidence about a null result \citep{hoenig2001abuse}. The study was not originally designed or powered to detect an effect as small as the one observed for the \(0.6\)-vs-\(0.9\) contrast. \Cref{tab:si:pairwise-comparisons} reports \(t\), Bonferroni-corrected \(p\), and Cohen's \(d\) for all three pairwise treatment comparisons on this analysed sample.

\begin{table}[h]
\centering
\small
\caption{\textbf{Pairwise treatment comparisons of participant-level mean Unsafe frequency, round \(t\geq2\) analysed sample.} Independent-samples \(t\)-tests (Bonferroni-corrected across three comparisons) and Cohen's \(d\), computed on the same round-\(t\geq2\), covariate-complete analysed sample as \Cref{tab:cluster_logit_unsafe} and \Cref{tab:si:summary-stats}. A separate set of tests on the raw, all-rounds per-participant sample (which additionally includes round-\(1\) decisions and participants without complete covariates) is reported in \Cref{fig:treatment_differences}A.}
\label{tab:si:pairwise-comparisons}
\begin{tabular}{lccccc}
\toprule
Comparison & \(N_1\) & \(N_2\) & \(t\) & \(p\) (Bonferroni) & Cohen's \(d\) \\
\midrule
\(p_r^{\max}=0.1\) vs.\ \(0.6\) & 97 & 105 & 2.424 & 0.0487 & 0.341 \\
\(p_r^{\max}=0.1\) vs.\ \(0.9\) & 97 & 136 & 2.422 & 0.0486 & 0.323 \\
\(p_r^{\max}=0.6\) vs.\ \(0.9\) & 105 & 136 & -0.206 & 1.000 & -0.027 \\
\bottomrule
\end{tabular}
\end{table}

\begin{figure}[H]
    \centering
    \includegraphics[width=0.6\linewidth]{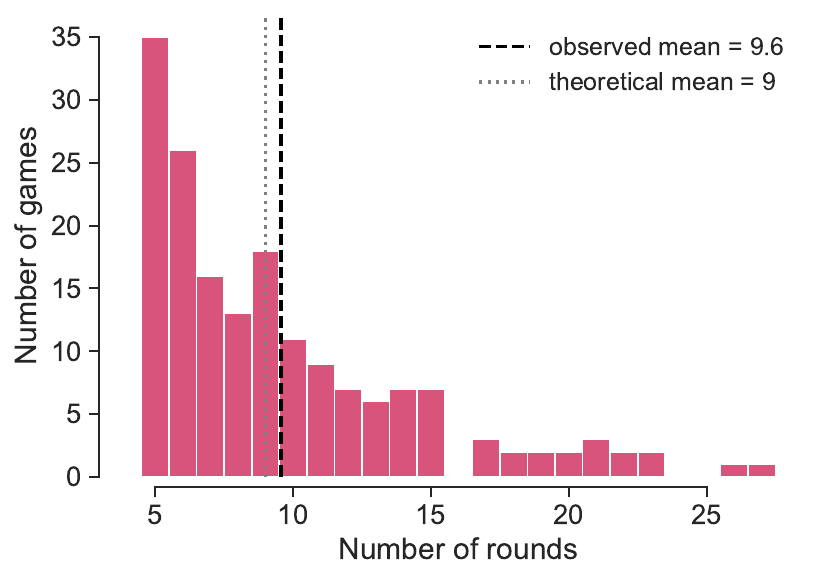}
    \caption{\textbf{Distribution of the number of rounds per game.} Realised number of rounds at which each of the \(N=173\) games in the experimental dataset ended, determined by the minimum of 5 rounds plus a 20\% per-round chance of stopping thereafter. The observed mean (\(9.56\) rounds, dashed black line) closely matches the correct theoretical expectation (\(9\) rounds, dotted grey line).}
    \label{fig:si:num_rounds}
\end{figure}

\paragraph{Mean number of rounds.} Both in the pre-registration and in the instructions to participants, we state that the main experimental task (the AI Race game) takes a minimum of 5 rounds, with a 20\% chance of ending after each subsequent round, making it on average 10 rounds. This is incorrect and a mistake on our part: the actual expected number of rounds under this stopping rule is \(5+1/0.2-1=9\), not 10. We only became aware of this arithmetic error while analysing the data, well after data collection was complete, so it should not have influenced participants' behaviour, and we do not consider it a form of deception. The error is also numerically minor since the geometric distribution governing the terminal round has high variance and a mode at 5, so the one-round discrepancy between the stated and the correct expected value is small relative to the spread of the distribution, and does not affect the main results or conclusions of the study. \Cref{fig:si:num_rounds} shows the realised distribution of the number of rounds per game in the experimental dataset (\(N=173\) games). The observed mean is \(9.56\) rounds, close to the correct theoretical expectation of 9.

\subsection{Data collection}
\label{si:sec:data-collection}

\subsubsection{Sample}
\label{si:sec:sample}

Participants were recruited through Prolific (cf.\ \Cref{si:sec:instructions-general}). A total of \(471\) participants were recruited across the three \(p_r^{\max}\) treatments (\(147\) for \(p_r^{\max}=0.1\), \(128\) for \(p_r^{\max}=0.6\), \(196\) for \(p_r^{\max}=0.9\)). Eligibility followed Prolific's standard platform pre-screening together with the comprehension-test and matching requirements described in \Cref{si:sec:exclusion}. After applying the exclusion criteria below, \(340\) participants (\(97\), \(104\), and \(139\) in the three treatments, respectively) completed the full experiment. The panel regressions reported in \Cref{tab:cluster_logit_unsafe} additionally restrict to round \(t\geq2\) and to participants with non-missing sex, age, nationality, and risk-preference covariates. Two of the 340 completers have a Prolific-flagged \texttt{DATA\_EXPIRED} nationality record and are excluded once nationality is added as a covariate, giving the final analysed sample of \(N=2{,}888\) observations, \(172\) pair clusters, and \(338\) participants (\(97\), \(105\), and \(136\) in the \(p_r^{\max}=0.1\), \(0.6\), and \(0.9\) treatments, respectively). These treatment-level counts are not simply the raw ``completed'' counts minus the two \texttt{DATA\_EXPIRED} exclusions: the \(p_r^{\max}=0.9\) count (\(139\to136\)) reflects exactly those two \texttt{DATA\_EXPIRED} participants, who completed the task but are dropped once nationality is required as a covariate; the \(p_r^{\max}=0.6\) count is instead one \emph{higher} than the raw completed figure (\(104\to105\)), because it additionally includes the one mid-race dropout described in \Cref{si:sec:attrition}, who did not satisfy the completion criteria in \Cref{tab:si:exclusions} but nonetheless contributed three valid round-\(t\geq2\) decisions and is therefore retained in the regression panel. Both \texttt{DATA\_EXPIRED} participants completed the AI-race task in full; they are retained in the publicly deposited dataset and excluded only from analyses that specifically require nationality as a covariate, consistent with the de-identification approach described in the Data Availability statement. A demographic and risk-preference breakdown of the analysed sample by treatment is given in \Cref{tab:si:summary-stats}.

\subsubsection{Exclusion criteria}
\label{si:sec:exclusion}

Participants could be excluded from the final analysed sample for the following reasons, described procedurally in \Cref{si:sec:instructions-task2,si:sec:survey}:
\begin{itemize}
    \item failing the comprehension test after five attempts;
    \item exceeding the 15-minute matching window without being paired with a competitor;
    \item exceeding the 2-minute decision time limit on any round;
    \item a paired competitor dropping out before the task was completed;
    \item dropping out prior to the risk-preference elicitation task (Task 1), before making any choice.
\end{itemize}

\Cref{tab:si:exclusions} reports the number of participants excluded for each reason, by treatment. The "Timed out (other)" column counts participants who timed out in any part of the experiment excluding the mid-race dropouts (e.g., timeouts in the comprehension test or in the risk-preference elicitation task). The "Not matched in time" column counts participants who exceeded the 15-minute matching window without being paired with a competitor.

\begin{table}[h]
\centering
\caption{\textbf{Pre-race exclusions, by treatment.} Counts are drawn from the raw Prolific/oTree completion-status field for all \(471\) recruited participants. Row totals are each 4/2/4 short of the recruited counts (\(147\), \(128\), \(196\); \Cref{si:sec:sample}) because the 10 participants affected by a mid-race decision timeout (5 pairs: 2 in \(p_r^{\max}=0.1\), 1 in \(0.6\), 2 in \(0.9\)) are excluded from this table and reported separately in \Cref{si:sec:attrition}.}
\label{tab:si:exclusions}
\begin{tabular}{lccccc}
\toprule
 & Failed & Dropped before & Timed out & Not matched & \\
\(p_r^{\max}\) & comprehension test & Task 1 & (other) & in time & Completed \\
\midrule
\(0.1\) & 7 & 0 & 39 & 0 & 97 \\
\(0.6\) & 4 & 0 & 15 & 3 & 104 \\
\(0.9\) & 16 & 4 & 33 & 0 & 139 \\
\midrule
Total & 27 & 4 & 87 & 3 & 340 \\
\bottomrule
\end{tabular}
\end{table}

Mid-race dropouts, in which a paired competitor drops out \emph{during} the AI-race task itself rather than before it, are reported separately in \Cref{si:sec:attrition}, since they affect participants who otherwise satisfy the completion criteria above.

\subsubsection{Attrition during the AI-race task}
\label{si:sec:attrition}

In five pairs (two in the \(p_r^{\max}=0.1\) treatment, one in \(p_r^{\max}=0.6\), and two in \(p_r^{\max}=0.9\)), one participant timed out on a decision during the AI-race task itself and was treated as having dropped out; their competitor's task ended at that point. These five dropouts and their five competitors are excluded from the ``completed'' counts in \Cref{tab:si:exclusions}, since their task was not completed in full.

Of these ten participants, only one (in the \(p_r^{\max}=0.6\) treatment) contributes any observations to the panel regressions in \Cref{tab:cluster_logit_unsafe}: this participant played four full rounds before dropping out, of which three fall in the round \(t\geq2\) analysis window used by \Cref{tab:cluster_logit_unsafe}. \Cref{tab:si:cluster_logit_unsafe_no_dropout} in \Cref{si:sec:data-analysis} reports \Cref{tab:cluster_logit_unsafe} recomputed after excluding this pair (\(N=2{,}885\) observations, \(171\) pair clusters): all coefficients are essentially unchanged, confirming that the one partially-completed pair does not drive the results.

\subsubsection{Data collection dates}
\label{si:sec:dates}

Data collection took place in two waves, corresponding to the original pre-registered treatments and the additional treatment described in \Cref{si:sec:prereg-changes}.
\begin{table}[h]
\centering
\caption{\textbf{Data collection waves.}}
\label{tab:si:collection-dates}
\begin{tabular}{lll}
\toprule
Wave & \(p_r^{\max}\) treatment(s) & Dates \\
\midrule
1 & \(0.6\) & 2024-06-25, 2024-07-15 \\
1 & \(0.9\) & 2024-08-15, 2024-10-31 \\
2 & \(0.1\) & 2024-11-11, 2025-07-30 \\
\bottomrule
\end{tabular}
\end{table}

\subsection{Experimental instructions and interface}
\label{si:sec:instructions}

Below we provide the instructions for the \(p_r^{\max}=0.6\) (60\%) treatment. The only difference between treatments is the maximum private risk value \(p_r^{\max}\) stated in the instructions.

\subsubsection{General instructions}
\label{si:sec:instructions-general}

You will participate in an experiment on decision-making conducted by the Teesside University and the Université Libre de Bruxelles (ULB). Your earnings will depend on your choices and the choices of other participants. Everyone receives the same instructions, which contain all the necessary information to complete the experiment, so please read carefully.\\

\noindent\textbf{Your privacy is guaranteed:} Your identity will be kept confidential, and results will be anonymous. For any questions, use the Prolific chat to contact the researcher.\\

\noindent\textbf{Earnings:} All your earnings during the experiment will be expressed in Experimental Currency Units (ECUs). Each ECUs is converted to POUNDS (£) with an exchange rate of 100 ECUs = £2, or 1 ECU = £0.02.\\

The experiment will last approximately 20 minutes, and it includes 2 tasks and a final survey. Your final earnings are calculated by summing a fixed amount (~£2 for 20 min) + a variable amount (paid as a bonus). The variable amount is calculated as the sum of earnings of the 2 tasks, and it will be a maximum of £5. We will only show you your final earnings at the end of the experiment.\\

\noindent\textbf{Notes:}

\begin{itemize}
    \item Completing the experiment is necessary to receive the bonus.
    \item Do not close the window, change tabs, refresh the page nor delete your cookies during the experiment to avoid being disqualified.
    \item If you exceed the maximum time assigned to a task, you will be considered to have abandoned the experiment and will not be paid.
\end{itemize}

Click Next to start Task 1 once you have read these instructions carefully. Once you click Next you cannot go back.

\subsubsection{Risk-preference elicitation task (Task 1)}
\label{si:sec:instructions-task1}

In this task, you will have to select one of six possible gambles. The six gambles are listed in the Table below.

\begin{itemize}
    \item You must select one gamble from the list.
    \item To select a gamble, click the radio button on the column "Your selection".
\end{itemize}

Each of these gambles has two possible outcomes, labelled as event A or event B, and both outcomes are equally likely to happen, with a 50\% chance (a coin flip). 

Once you select a gamble and click the button Next, the computer will flip a virtual coin to choose one of the two possible events (A or B). You will only be able to observe this outcome at the end of the experiment.

The amount of money or reward you receive will be determined by:

\begin{itemize}
    \item The gamble you choose, and 
    \item which outcome occurs.
\end{itemize}

For example, if you select Gamble 3 and event A occurs, you will earn 20 ECUs. If event B occurs, you will earn 44 ECUs.

\begin{figure}[H]
    \centering
    \includegraphics[width=1.0\linewidth]{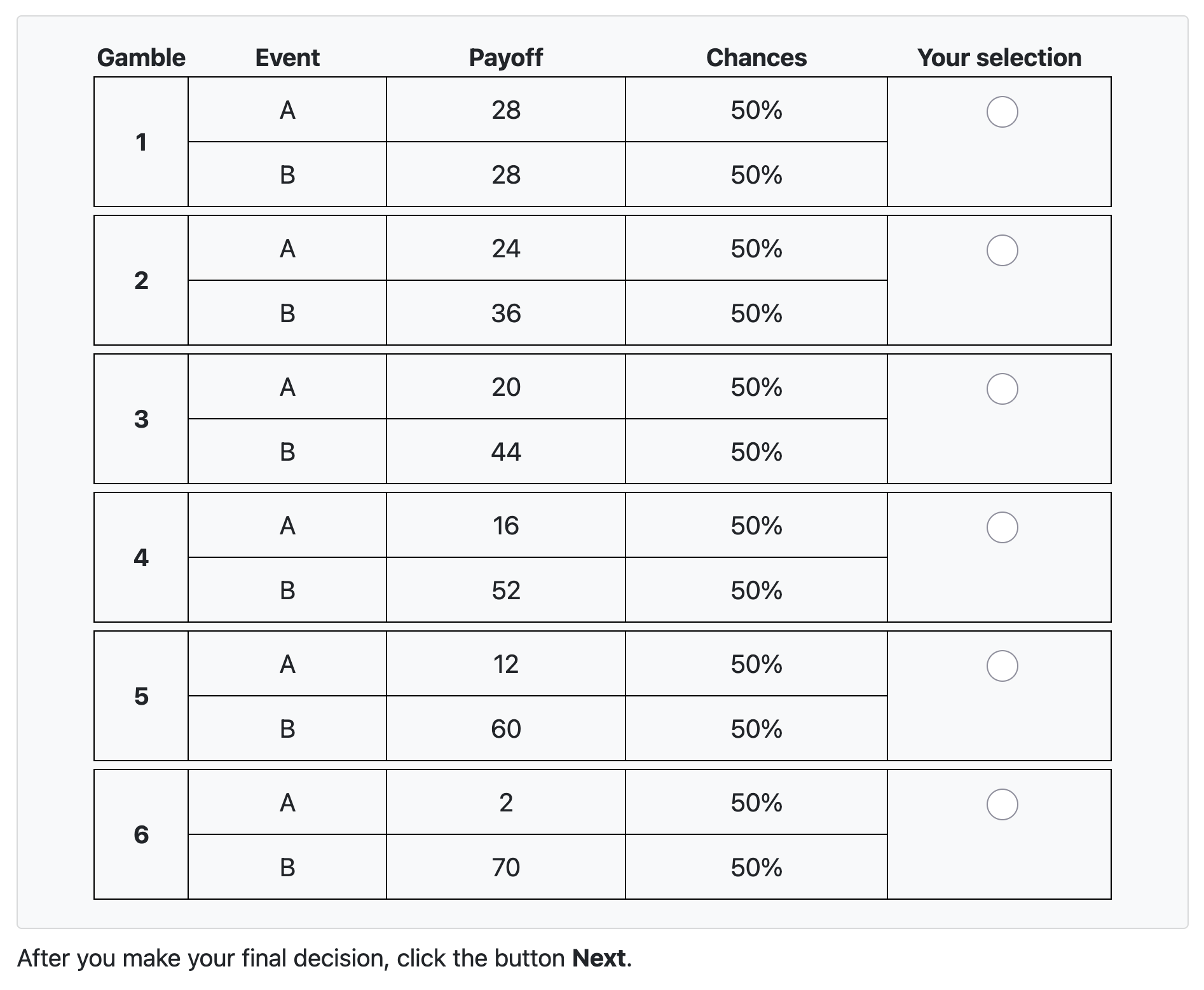}
    \caption{Visual representation of the choice in Task 1.}
    \label{fig:si:instructions_task1}
\end{figure}

Click Next when you have made your decision. Once you click Next, you cannot go back.

\subsubsection{AI-race task instructions (Task 2)}
\label{si:sec:instructions-task2}

\paragraph{Instructions 1/4: General Description}

\begin{itemize}
    \item \textbf{Pairing:} You will be randomly matched with another participant, referred to as the competitor.
    \item \textbf{Rounds:} The game lasts a minimum of 5 rounds, and, on average, 10 rounds.
    \item \textbf{Anonymity:} Your identity will be kept private. Both players receive the same instructions and see the same information on their screens.
\end{itemize}

\noindent In this task, you and the participant you have been matched with (your competitor), are asked to take on the role of a stakeholder, each representing a different company participating in a technological race. Both of you will be competing for technological supremacy. The faster you develop your company’s technology the closer you will get to technological supremacy. In summary:

\begin{itemize}
    \item \textbf{Role:} You and the competitor represent different companies in a technological race. Throughout several rounds you can develop your company’s technology in a SAFE or UNSAFE manner. After the final round, whoever made the most development steps achieves technological supremacy.
    \item \textbf{Steps per round:}
    \begin{enumerate}
        \item Both players choose between 2 options (act SAFE/UNSAFE) to advance in the race.
        \item Payoff depends on both players' choices.
        \item After round 5, a virtual dice roll decides if the game continues.
    \end{enumerate}
\end{itemize}

\paragraph{Instructions 2/4: Calculate Your Earnings}

\begin{itemize}
    \item \textbf{Choices:}
    \begin{itemize}
        \item Act SAFE and you will take 1 step towards technological supremacy. This option does not entail risk.
        \item Act UNSAFE and you will take 1.5 steps closer to technological supremacy. This option entails risk (explained in the next page).
    \end{itemize}
    \item \textbf{Winning:} The participant with the most steps at the end wins the race.
    \item \textbf{Round Payoff Calculation (see payoff matrix):}
    \begin{itemize}
        \item Both act SAFE: Each gets 1 ECU.
        \item You act SAFE, the competitor acts UNSAFE: You get 0.6 ECUs, the competitor gets 2.4 ECUs.
        \item You act UNSAFE, the competitor acts SAFE: You get 2.4 ECUs, the competitor gets 0.6 ECUs.
        \item Both act UNSAFE: Each gets 2 ECUs.
    \end{itemize}
    \item \textbf{Final Payoff before risk of setback:} Sum of all round payoffs plus a 100 ECUs bonus for the winner. If tied, each gets 50 ECUs.
\end{itemize}

\paragraph{Instructions 3/4: Final Round and Risk of Setback}

\begin{itemize}
    \item \textbf{Ending the Game:} After round 5, each round has a 20\% chance of ending the game.
    \item \textbf{Risk of personal setback:} The more UNSAFE actions you take, the more risk that you will suffer a personal setback if you win the race. The risk is calculated at the end of the game as (Number of UNSAFE actions / Total actions) * 60\%. This means that the maximum risk of personal setback is 60\% (if you always act UNSAFE) and the minimum is 0\% (if you never act UNSAFE). Your personal risk is displayed on your screen at each round.
    \item \textbf{Who may suffer a personal setback:} only the race winner(s) may suffer a personal setback.
    \item \textbf{Effect of personal setback:} If you win the race, and you suffer a personal setback, you lose all your earnings, including the 100 ECUs bonus.
\end{itemize}

\paragraph{Instructions 4/4: Important Notes}

\begin{itemize}
    \item \textbf{Comprehension Test:} You must pass a short comprehension test to proceed. You have 5 attempts. If you fail 5 times, the experiment will be terminated.
    \item \textbf{Waiting:} You may wait up to 15 minutes to be matched with another participant. During this time, leaving the tab or browser will remove you from the experiment. If we cannot match you with another participant in time, the experiment will be terminated.
    \item \textbf{Decision Time:} Each decision has a 2-minute limit. Timing out results in removal from the experiment.
    \item \textbf{Dropouts:} If the competitor drops out, the task is cancelled, and you will be paid for the completed task and your earnings in the current task before it was cancelled.
\end{itemize}

Click "Next" to begin the comprehension test. Once you click Next, you cannot go back.

\begin{figure}[H]
    \centering
    \includegraphics[width=1.0\linewidth]{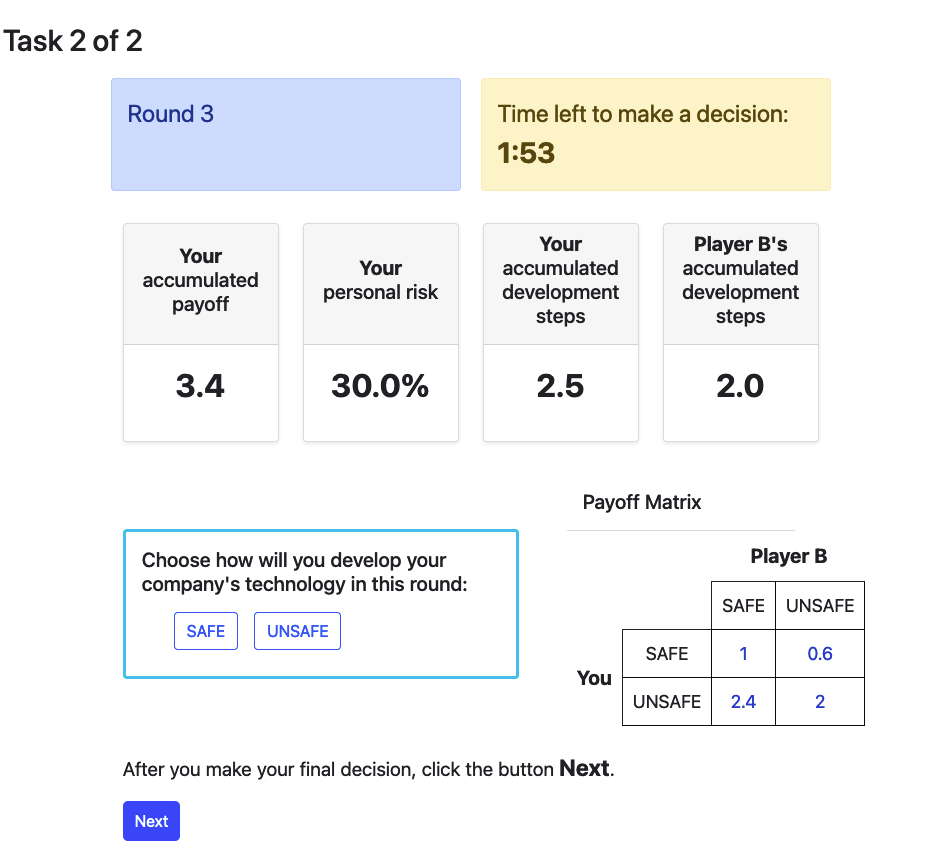}
    \caption{Visual representation of the choice in Task 2.}
    \label{fig:si:instructions_task2_decision}
\end{figure}

\begin{figure}[H]
    \centering
    \includegraphics[width=1.0\linewidth]{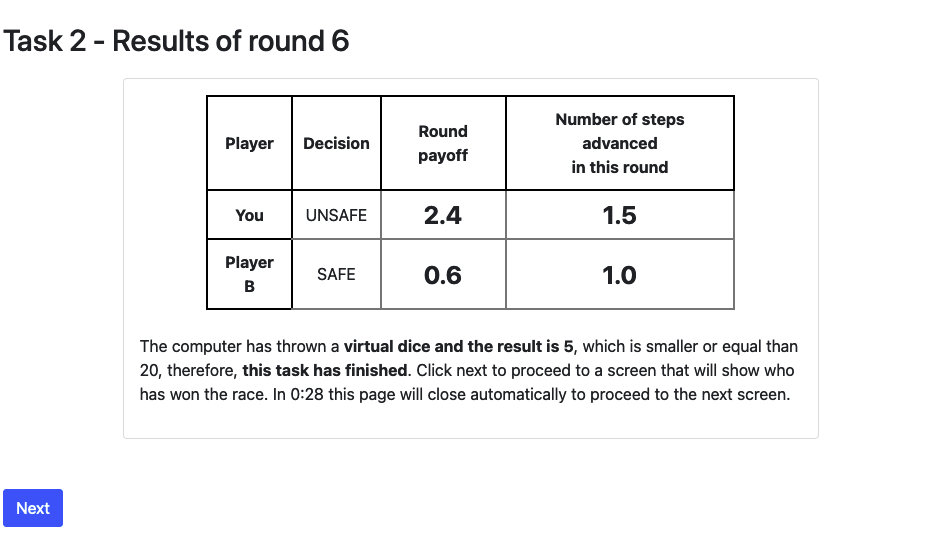}
    \caption{Visual representation of round results in Task 2.}
    \label{fig:si:instructions_task2_info}
\end{figure}

\subsubsection{Post-experiment survey}
\label{si:sec:survey}

At the end of the experiments, participants were asked to answer the following short survey:

\begin{enumerate}
    \item In the last task, what was your reasoning behind choosing SAFE/UNSAFE? (open text)
    \item Did your opponent's choices affect your own? (open text)
    \item Did you expect your competitor to choose SAFE/UNSAFE more often? Why? (open text)
    \item Are there any comments or feedback you would like to provide for this experiment? (open text)
    \item Were you familiar with these types of experiments? Yes/No
    \item Choose the point on the scale that matches how much you agree with the following statements (5-point likert scale).
    \begin{enumerate}
        \item You are familiar with Game Theory
        \item You are familiar with Behavioural experiments
        \item You are familiar with the game you played in task 2
    \end{enumerate}
\end{enumerate}

\section{Supplementary data analysis}
\label{si:sec:data-analysis}

\subsection{Summary statistics}
\label{si:sec:summary-stats}

\Cref{tab:si:summary-stats} reports summary statistics for the analysed sample: the number of participants and pairs, sex balance, age, nationality balance, mean frequency of Unsafe choices \(\phi_U\), and Eckel--Grossman risk-preference (gamble-choice, \(0\)--\(5\)) balance, by \(p_r^{\max}\) treatment. Nationality is collapsed to South Africa and Poland, the two dominant recruitment pools (\(44\%\) and \(11\%\) of participants, respectively), versus all other nationalities, since the full set of \(\sim\!46\) distinct nationalities is too sparse to report individually.

\begin{table}[h]
\centering
\small
\caption{\textbf{Summary statistics of the analysed sample.} N participants and N pairs, sex balance, age (mean, SD), nationality balance, mean frequency of Unsafe choices (\(\phi_U\)), and Eckel--Grossman risk-preference (gamble choice, 0--5) balance, by maximum private-risk treatment. Computed on the same analysed sample as \Cref{tab:cluster_logit_unsafe} (round \(t\geq2\), N=2,888 observations, 172 pair clusters, 338 participants).}
\label{tab:si:summary-stats}
\begin{tabular}{lcccc}
\toprule
Statistic & \(p_r^{\max}=0.1\) & \(p_r^{\max}=0.6\) & \(p_r^{\max}=0.9\) & Total \\
\midrule
N participants & 97 & 105 & 136 & 338 \\
N pairs & 49 & 53 & 70 & 172 \\
Female & 50 & 46 & 72 & 168 \\
Male & 47 & 58 & 63 & 168 \\
Prefer not to say & 0 & 1 & 1 & 2 \\
Mean age (SD) & 29.0 (9.8) & 27.7 (6.2) & 28.8 (9.5) & 28.5 (8.7) \\
Nationality: South Africa & 67 & 24 & 58 & 149 \\
Nationality: Poland & 7 & 14 & 16 & 37 \\
Nationality: Other & 23 & 67 & 62 & 152 \\
Mean \(\phi_U\) & 0.640 & 0.558 & 0.564 & 0.584 \\
Mean gamble choice & 1.85 & 2.22 & 2.37 & 2.17 \\
Modal gamble choice & 0 & 2 & 2 & 2 \\
\bottomrule
\end{tabular}
\end{table}

\subsection{Additional covariates: demographics and risk preference}
\label{si:sec:data-analysis-demographics}
\label{si:sec:data-analysis-risk}

\Cref{tab:cluster_logit_unsafe} reports the treatment, lagged-action, and race-position coefficients from the cluster-robust panel logistic regressions of Unsafe choices; every model in that table also includes demographic covariates for participant sex, age, and nationality, together with a risk-preference covariate based on the \cite{eckel2008forecastingRiskAttitudes} elicitation task described in \Cref{si:sec:instructions-task1}, whose coefficients are omitted there to avoid cluttering the main table. \Cref{tab:si:panel-covariates} reports these omitted coefficients for the corresponding model specifications. Risk preference is entered as a continuous score (the participant's gamble choice, \(0\)--\(5\)): the six gambles have monotonically increasing payoff risk (standard deviation \(0,6,12,18,24,34\)) by design, following the same convention \cite{eckel2008forecastingRiskAttitudes} themselves use when treating gamble choice as a numeric risk-tolerance ranking. The interaction rows are estimated from a parallel set of models that replace the additive risk-preference term with its interaction with the treatment dummies (reference \(p_r^{\max}=0.1\)). None of these covariates significantly predicts Unsafe choices.

The point estimate on the risk-preference coefficient is consistently negative across all six models, which would, taken at face value, suggest that participants who chose a higher-payoff-risk gamble (i.e.\ who are less risk averse) were slightly \emph{less} likely to choose Unsafe in the race, the opposite of a naive expectation that less risk-averse participants take more risks generally. However, all risk-preference coefficients, in both the additive and interaction specifications, are small in magnitude and not statistically distinguishable from zero (\(p>0.1\) throughout; \Cref{tab:si:panel-covariates}). We therefore conclude that the Eckel--Grossman gamble choice does not predict Unsafe behaviour in our experiment, consistent with the analysis in \Cref{si:sec:prereg-analyses}.

\begin{table}[h]
\centering
\small
\caption{\textbf{Demographic- and risk-preference-covariate coefficients from the panel logistic regressions of Unsafe decisions.} Coefficients other than those listed here are identical to the corresponding column of \Cref{tab:cluster_logit_unsafe} and are not repeated. Standard errors clustered at the pair level are reported in parentheses. Female is the reference sex category, so the Sex coefficient reports the change in log-odds of choosing Unsafe for male participants relative to female participants. Age is centred on its sample mean. Nationality is South Africa or Poland (the two dominant recruitment pools, \(44\%\) and \(11\%\) of participants, respectively) relative to all other nationalities. \(^{\dagger}p<0.1\), \(^{*}p<0.05\), \(^{**}p<0.01\), \(^{***}p<0.001\).}
\label{tab:si:panel-covariates}
\begin{tabular}{lcccccc}
\toprule
 & \multicolumn{3}{c}{Without \(a_i^1\)}  & \multicolumn{3}{c}{With \(a_i^1\)} \\
\cmidrule(lr){2-4}\cmidrule(lr){5-7}
 & (1) & (2) & (3) & (4) & (5) & (6) \\
\midrule
Sex (Male) & 0.110 & 0.084 & 0.083 & 0.055 & 0.049 & 0.046 \\
 & (0.120) & (0.108) & (0.109) & (0.122) & (0.113) & (0.114) \\
Age (centred) & 0.006 & 0.005 & 0.005 & 0.006 & 0.004 & 0.005 \\
 & (0.007) & (0.006) & (0.006) & (0.007) & (0.007) & (0.006) \\
Nationality (South Africa) & 0.152 & 0.130 & 0.112 & 0.154 & 0.134 & 0.117 \\
 & (0.148) & (0.129) & (0.127) & (0.147) & (0.131) & (0.129) \\
Nationality (Poland) & -0.137 & -0.122 & -0.098 & -0.130 & -0.128 & -0.104 \\
 & (0.182) & (0.164) & (0.165) & (0.195) & (0.173) & (0.174) \\
Risk preference & -0.012 & -0.017 & -0.013 & -0.024 & -0.027 & -0.023 \\
 & (0.033) & (0.029) & (0.030) & (0.033) & (0.030) & (0.031) \\
Risk preference \(\times\ p_r^{\max}=0.6\) & -0.013 & -0.019 & -0.009 & -0.005 & -0.016 & -0.006 \\
 & (0.079) & (0.071) & (0.071) & (0.079) & (0.072) & (0.074) \\
Risk preference \(\times\ p_r^{\max}=0.9\) & -0.114 & -0.109 & -0.110 & -0.095 & -0.101 & -0.101 \\
 & (0.085) & (0.075) & (0.075) & (0.084) & (0.076) & (0.076) \\
\bottomrule
\end{tabular}
\end{table}

\subsection{Robustness: excluding the mid-race dropout}
\label{si:sec:data-analysis-robustness}

\Cref{tab:si:cluster_logit_unsafe_no_dropout} repeats \Cref{tab:cluster_logit_unsafe} after excluding the one pair affected by a mid-race decision timeout (\Cref{si:sec:attrition}). All coefficients are essentially unchanged, confirming that this single partially-completed pair does not drive the reported results.

\begin{table}[h]
\centering
\small
\caption{\textbf{Robustness: cluster-robust panel logistic regression of Unsafe decisions excluding the mid-race dropout.} Identical specification to \Cref{tab:cluster_logit_unsafe}, with the one pair affected by a mid-race decision timeout removed. Standard errors clustered at the pair level are reported in parentheses. \(^{\dagger}p<0.1\), \(^{*}p<0.05\), \(^{**}p<0.01\), \(^{***}p<0.001\).}
\label{tab:si:cluster_logit_unsafe_no_dropout}
\begin{tabular}{lcccccc}
\toprule
 & \multicolumn{3}{c}{Without \(a_i^1\)}  & \multicolumn{3}{c}{With \(a_i^1\)} \\
\cmidrule(lr){2-4}\cmidrule(lr){5-7}
 & (1) & (2) & (3) & (4) & (5) & (6) \\
\midrule
\(p_r^{\max}=0.6\) & -0.180 & -0.139 & -0.149 & -0.151 & -0.119 & -0.128 \\
 & (0.211) & (0.180) & (0.179) & (0.210) & (0.182) & (0.181) \\
\(p_r^{\max}=0.9\) & -0.245 & -0.190 & -0.194 & -0.241 & -0.188 & -0.191 \\
 & (0.183) & (0.155) & (0.155) & (0.180) & (0.155) & (0.155) \\
\(a_i^1\) &  &  &  & 0.293\(^{*}\) & 0.209\(^{\dagger}\) & 0.219\(^{\dagger}\) \\
 &  &  &  & (0.117) & (0.118) & (0.116) \\
\(a_i^{t-1}\) &  & 0.022 & -0.170 &  & 0.006 & -0.191 \\
 &  & (0.128) & (0.196) &  & (0.128) & (0.192) \\
\(a_{-i}^{t-1}\) &  & 0.863\(^{***}\) & 0.644\(^{**}\) &  & 0.832\(^{***}\) & 0.611\(^{**}\) \\
 &  & (0.123) & (0.197) &  & (0.121) & (0.192) \\
\(\Delta S_{t-1}\) &  & 0.107\(^{\dagger}\) & -0.238 &  & 0.066 & -0.297\(^{*}\) \\
 &  & (0.057) & (0.150) &  & (0.067) & (0.149) \\
\(a_i^{t-1}\times a_{-i}^{t-1}\) &  &  & 0.389 &  &  & 0.393\(^{\dagger}\) \\
 &  &  & (0.240) &  &  & (0.239) \\
\(a_i^{t-1}\times \Delta S_{t-1}\) &  &  & 0.448\(^{*}\) &  &  & 0.466\(^{*}\) \\
 &  &  & (0.185) &  &  & (0.182) \\
\(a_{-i}^{t-1}\times \Delta S_{t-1}\) &  &  & 0.206 &  &  & 0.223 \\
 &  &  & (0.213) &  &  & (0.214) \\
\(a_i^{t-1}\times a_{-i}^{t-1}\times \Delta S_{t-1}\) &  &  & -0.167 &  &  & -0.189 \\
 &  &  & (0.211) &  &  & (0.210) \\
Demographic covariates & Yes & Yes & Yes & Yes & Yes & Yes \\
Risk-preference covariates & Yes & Yes & Yes & Yes & Yes & Yes \\
Observations & 2,885 & 2,885 & 2,885 & 2,885 & 2,885 & 2,885 \\
Pair clusters & 171 & 171 & 171 & 171 & 171 & 171 \\
Pseudo \(R^2\) & 0.006 & 0.035 & 0.038 & 0.009 & 0.037 & 0.040 \\
\bottomrule
\end{tabular}
\end{table}

\section{Supplementary model analysis}
\label{si:sec:model-analysis}

\subsection{Payoff calculations}
\label{si:expected-payoff-matrix}

\subsubsection{Two-player indefinitely repeated AI race game}

We consider a two-player indefinitely repeated AI race game with a discount factor $\delta$. In each round, each player chooses one of two actions:
\[
S \quad \text{(Safe)}, \qquad U \quad \text{(Unsafe)}.
\]
The action profile determines both the round payoffs and the progress each player makes in the race.

\subsubsection{Parameters}

The model uses the following parameters:
\[
b = 4, \qquad B = 100, \qquad c = 1,
\]
\[
s_U = 1.5, \qquad s_S = 1,
\]
where \(b\) is the per-round prize from the strategic interaction, \(B\) is the final race prize, \(c\) is the cost of playing Safe, and \(s_S\) and \(s_U\) are the race steps obtained from actions \(S\) and \(U\), respectively.

The game lasts for at least five rounds. After round \(5\), and after every subsequently completed round, the game ends with probability \(p\in(0,1)\) ($\delta=1-p$). Thus the total number of rounds is
\[
W = 5 + G,
\qquad
G \sim \mathrm{Geom}(p)-1,
\]
where \(\mathrm{Geom}(p)\) denotes the geometric distribution on \(\{1,2,\dots\}\), so that
\[
\Pr(W=w)=p(1-p)^{w-5},
\qquad w=5,6,7,\dots.
\]
This specialises to the experimental stopping rule defined in Methods (\Cref{sec:methods}) at \(p=0.2\) (\(\delta=0.8\)), for which \(\mathbb{E}[W]=9\).

\subsubsection{Stage-game payoffs}

Let the stage-game payoff matrix be
\[
\begin{array}{c|cc}
 & S & U\\
\hline
S & \pi_{11} & \pi_{12}\\
U & \pi_{21} & \pi_{22}
\end{array}
\]
with
\[
\pi_{11} = -c + \frac{b}{2},
\qquad
\pi_{12} = -c + \frac{b}{s_U+1},
\]
\[
\pi_{21} = \frac{s_U b}{s_U+1},
\qquad
\pi_{22} = \frac{b}{2}.
\]
Under the parameter values above, this reduces to the numeric payoff matrix already given in Methods (\Cref{sec:methods}): \(\pi_{11}=1\), \(\pi_{12}=0.6\), \(\pi_{21}=2.4\), \(\pi_{22}=2\).

\subsubsection{Strategies and the distance-dependent approximation}

Recall the four strategies defined in Methods (\Cref{sec:methods}): Always Safe (AS), Always Unsafe (AU), Conditionally Safe (CS), and Conditionally Antisocial Safe (CAS).

The experimental data suggest that some participants condition their propensity to play the Unsafe action \(U\) on their relative position in the race. In particular, the probability of Unsafe play appears to increase when a participant falls sufficiently behind the opponent, consistent with a distance-dependent or threshold-like behavioural rule. Such behaviour is more complex than the deterministic memory-one strategies included in our reduced analytical model.

To keep the strategy space tractable, we do not model this distance-dependent rule explicitly. Instead, we interpret the conditional strategies CS and CAS as approximations to different components of the empirically observed behaviour. Strategy CS captures a comparatively conservative conditional response, while CAS captures a more aggressive response that places greater weight on race position and catching up. In this sense, both strategies can be viewed as reduced-form proxies for a richer behavioural rule in which the probability of choosing \(U\) depends on the current race distance.

The purpose of the reduced strategy set is to capture the main qualitative patterns of conditional play while preserving analytical tractability, not to fully reproduce individual behaviour.

\subsubsection{Race prize and private risk}

Let the total number of Unsafe actions played by player \(i\) up to round \(W\) be denoted by \(n_i(W)\). Total race progress is then
\[
R_i(W)=Ws_S + \bigl(s_U-s_S\bigr)n_i(W).
\]
A player wins the race if its final progress is at least as large as that of the opponent. Ties are treated as joint wins and split the final prize equally.

A player is exposed to private risk only if it wins or ties the race. Let \(p_r^{\max}\) denote the maximum private risk. If a winning or tying player has played Unsafe in a fraction \(n_i(W)/W\) of rounds, then its effective private risk is
\[
p_r(W)=p_r^{\max}\frac{n_i(W)}{W}.
\]
Accordingly, if player \(i\) wins outright, its final payoff is multiplied by
\[
1-p_r^{\max}\frac{n_i(W)}{W},
\]
whereas in the case of a tie the same factor applies to its share \(B/2\). If a player does not win or tie the race, it receives no final prize and bears no private risk.

\subsubsection{Expected payoffs}
\label{si:sec:expected-payoffs}

Let \(\Pi_{X,Y}(W)\) denote the realised total payoff (round payoffs plus race prize, net of private risk) obtained by strategy \(X\) against strategy \(Y\) conditional on the game lasting exactly \(W\) rounds. Since the duration \(W\) is itself stochastic, we compute expected payoffs as
\[
\overline{\Pi}_{X,Y}
=
\mathbb{E}\bigl[\Pi_{X,Y}(W)\bigr]
=
\sum_{w=5}^{\infty}p(1-p)^{w-5}\Pi_{X,Y}(w).
\]

When neither player uses a conditional strategy, i.e.\ for AS against AS or AU, and AU against AS or AU, the realised round-by-round payoff is affine in \(W\), so \(\overline{\Pi}_{X,Y}\) is obtained directly by substituting \(\mathbb{E}[W]\):
\[
\overline{\Pi}_{AS,AS}=\frac{B}{2}+\mathbb{E}[W]\,\pi_{11},
\qquad
\overline{\Pi}_{AS,AU}=\mathbb{E}[W]\,\pi_{12},
\]
\[
\overline{\Pi}_{AU,AS}=\left(1-p_r^{\max}\right)\bigl(B+\mathbb{E}[W]\,\pi_{21}\bigr),
\qquad
\overline{\Pi}_{AU,AU}=\left(1-p_r^{\max}\right)\left(\frac{B}{2}+\mathbb{E}[W]\,\pi_{22}\right).
\]

Once one or both players use a conditional strategy (CS or CAS), realised play depends on the actual sequence of actions taken, and hence on the parity of \(W\) and, for the winning player, on the realised fraction of Unsafe actions \(n_i(W)/W\) that enters the private-risk-adjusted prize. These payoffs are not affine in \(W\) and, in general, do not reduce to a compact closed form. We therefore compute \(\overline{\Pi}_{X,Y}\) for every matchup involving CS and/or CAS by Monte Carlo simulation: for each such ordered strategy pair, we simulate \(10^4\) independent realisations of the race. The sample mean over the \(10^4\) realisations is used as the estimate of \(\overline{\Pi}_{X,Y}\) in the payoff matrix analysed in \Cref{fig:model_predictions}.

\subsection{Relationship to the repeated Prisoner's Dilemma}
\label{si:sec:pd-relationship}

The AI-race game shares a key feature with the repeated Prisoner's Dilemma (PD): in both games, the socially costly action strictly dominates the cooperative one in every round. Writing the stage-game payoffs (\Cref{sec:methods}) in the standard PD notation -- temptation \(T=\pi_{21}\), reward \(R=\pi_{11}\), punishment \(P=\pi_{22}\), and sucker's payoff \(S=\pi_{12}\) -- our numeric values give \(T=2.4\), \(P=2\), \(R=1\), \(S=0.6\), so that \(T>P>R>S\). This is the ordering of a \emph{Deadlock} game rather than a PD, which additionally requires \(R>P\). Unsafe strictly dominates Safe each round exactly as Defect dominates Cooperate in a PD, but mutual Unsafe play is not itself a socially costly outcome at the stage-game level: absent any terminal risk, two players who always play Unsafe are jointly better off than two players who always play Safe.

The accumulated private risk changes this. Using the closed-form total payoffs already derived above, \(\overline{\Pi}_{AS,AS}=B/2+\mathbb{E}[W]\pi_{11}=59\) is independent of \(p_r^{\max}\), whereas \(\overline{\Pi}_{AU,AU}=(1-p_r^{\max})(B/2+\mathbb{E}[W]\pi_{22})=(1-p_r^{\max})\cdot68\) decreases with it. Mutual Safe therefore yields a higher \emph{total} expected payoff than mutual Unsafe once
\[
p_r^{\max} > p_r^{\max*} = 1-\frac{B/2+\mathbb{E}[W]\pi_{11}}{B/2+\mathbb{E}[W]\pi_{22}} = 1-\frac{59}{68}\approx0.132,
\]
which holds for the \(p_r^{\max}=0.6\) and \(0.9\) treatments (and is nearly reached already at \(p_r^{\max}=0.1\)). The repeated structure of the race therefore turns an otherwise non-dilemma stage game into a social dilemma at the level of total expected payoffs, even though Unsafe remains the tempting round-by-round choice throughout.

\paragraph{Nash equilibria of the unconditional strategies.} Restricting attention to the two unconditional strategies AS and AU, and reusing \(\overline{\Pi}_{AS,AU}=\mathbb{E}[W]\pi_{12}=5.4\) and \(\overline{\Pi}_{AU,AS}=(1-p_r^{\max})(B+\mathbb{E}[W]\pi_{21})=(1-p_r^{\max})\cdot121.6\) from above, \((AU,AU)\) is a Nash equilibrium of this restricted game whenever \(\overline{\Pi}_{AU,AU}\ge\overline{\Pi}_{AS,AU}\), i.e. \(p_r^{\max}\le1-5.4/68\approx0.921\), and \((AS,AS)\) is a Nash equilibrium whenever \(\overline{\Pi}_{AS,AS}\ge\overline{\Pi}_{AU,AS}\), i.e. \(p_r^{\max}\ge1-59/121.6\approx0.515\). For \(p_r^{\max}<0.515\), \((AU,AU)\) is the unique equilibrium; for \(0.515\le p_r^{\max}\le0.921\), both \((AS,AS)\) and \((AU,AU)\) are Nash equilibria of this two-strategy game, a coordination structure in which both are strict equilibria and therefore both correspond to stable rest points of the associated replicator dynamics, separated by an unstable interior mixture; for \(p_r^{\max}>0.921\), \((AS,AS)\) is the unique equilibrium. So even without any conditional strategy, whether Safe is ever a best response to itself already depends sharply on \(p_r^{\max}\).

\paragraph{Nash equilibria of the full four-strategy game.} Adding the conditional strategies CS and CAS, we identify pure-strategy Nash equilibria by exhaustive best-response comparison on the same expected-payoff matrix used to build \Cref{fig:model_predictions} and \Cref{fig:si:strategy-frequencies} (\Cref{tab:si:pd-nash}). Two structural facts drive the result. First, CAS is behaviourally indistinguishable from AU whenever matched against an opponent that plays Unsafe from the first round onward: both then play Unsafe in every round from round \(2\) on, differing only in the payoff-irrelevant first round, so \((AU,AU)\), \((AU,CAS)\), \((CAS,AU)\), and \((CAS,CAS)\) are simultaneously either all Nash equilibria or none. Second, Always Safe is never itself a Nash equilibrium once conditional strategies are available: a CAS mutant that plays Unsafe once and Safe thereafter overtakes an Always-Safe resident by exactly half a race step every game and so wins the race almost risk-free, which strictly dominates AS at every \(p_r^{\max}\) we consider. Conditionally Safe, in contrast, becomes the unique equilibrium at high private risk: matched against an Unsafe opponent, CS plays Safe in round \(1\) and Unsafe thereafter, which leaves it permanently half a step behind and therefore never exposed to the terminal private-risk lottery, so once that lottery is costly enough, deliberately staying behind is safer than winning.

\begin{table}[h]
\centering
\small
\caption{\textbf{Pure-strategy Nash equilibria of the reduced four-strategy game, by treatment.} Computed from the same expected-payoff matrix used for \Cref{fig:model_predictions}. AU and CAS are payoff-equivalent whenever matched with each other (see text), so the equilibrium set at low and intermediate \(p_r^{\max}\) includes all four combinations of the two; AS is never part of an equilibrium at any treatment.}
\label{tab:si:pd-nash}
\begin{tabular}{lll}
\toprule
\(p_r^{\max}\) & Symmetric Nash equilibria & Is Always Safe an equilibrium? \\
\midrule
0.1 & AU, CAS (equivalent) & No \\
0.6 & AU, CAS (equivalent) & No \\
0.9 & CS (unique) & No \\
\bottomrule
\end{tabular}
\end{table}

At \(p_r^{\max}=0.1\) and \(0.6\), the AU/CAS equilibrium is the only one; the \(p_r^{\max}=0.6\) treatment lies close to, but just below, the point at which the population-level phase transition toward Conditionally Safe already reported in \Cref{si:sec:strategy-distribution} (\Cref{fig:si:strategy-frequencies}) begins. At \(p_r^{\max}=0.9\), Conditionally Safe is the unique equilibrium. Always Safe -- unconditional cooperation regardless of what the opponent does -- is therefore never itself a Nash equilibrium of the reduced game at any of the three treatments we study. What becomes favoured at high private risk is instead a \emph{conditional} strategy that starts Safe and continues to mirror the opponent.

\paragraph{Relation to strategies studied in the repeated Prisoner's Dilemma literature.} CS and CAS are memory-one reactive strategies: each copies the opponent's previous action after a fixed first move, which makes them instances of Tit-for-Tat (starting Safe or Unsafe, respectively). Reactive, TFT-like strategies of this kind are among the strategies most frequently inferred from human play in indefinitely and finitely repeated PD experiments \citep{dal2018determinants,monteroPorras2022inferringStrategies}, alongside unconditional cooperation and defection. The strong, robust effect of the opponent's previous action found in \Cref{tab:cluster_logit_unsafe} is consistent with this broader literature, and supports reading our Conditionally Safe and Conditionally Antisocial Safe strategies as race-specific cases of reciprocal play documented well beyond this particular game, rather than as evidence of a behavioural state unique to competitive racing.

\subsection{Strategy distribution as a function of \texorpdfstring{$\beta$}{beta}, \texorpdfstring{$p_r^{\max}$}{pr max}, and \texorpdfstring{$\mu$}{mu}}
\label{si:sec:strategy-distribution}

\begin{figure}[H]
    \centering
    \includegraphics[width=1.0\linewidth]{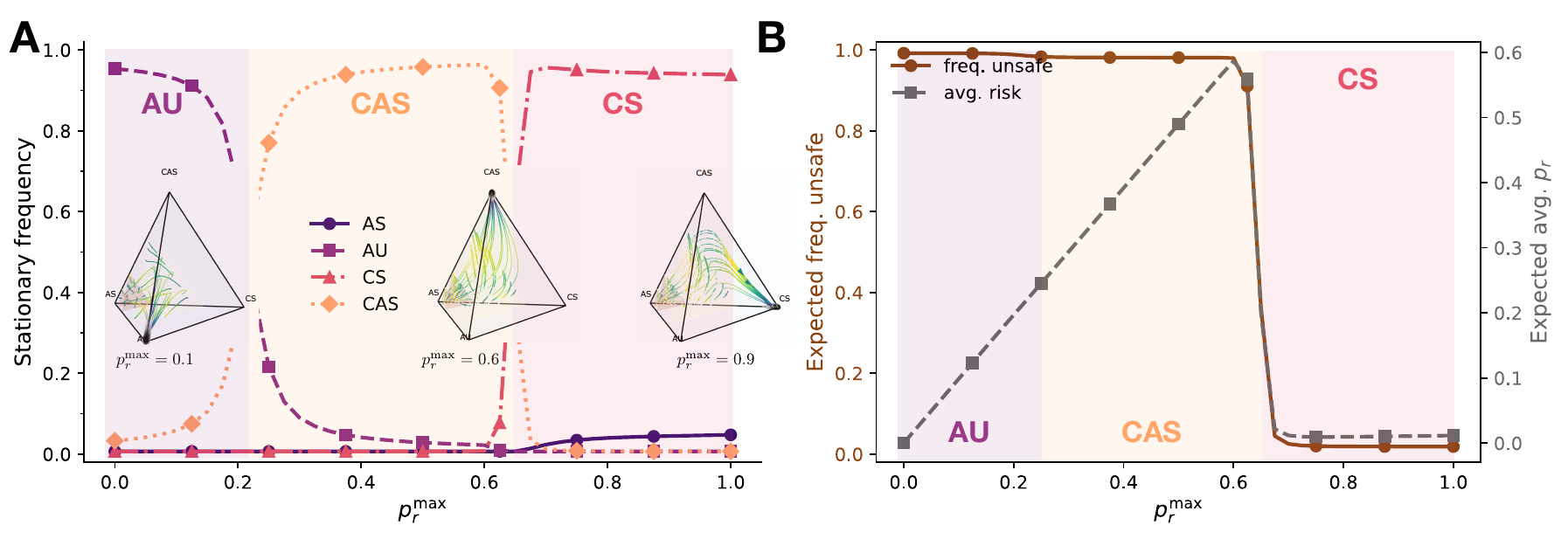}
    \caption{Frequency of strategies and expected unsafe behaviour in function of \(p_r^{\max}\). (A) Frequency of strategies in function of \(p_r^{\max}\). We observe phase transitions at $p_r^{\max}=0.2$ and $0.6$. For $p_r^{\max}<0.2$ AU dominates, while for $0.2<p_r^{\max}<0.6$ CAS dominates, and for $p_r^{\max}>0.6$ CS dominates. (B) Expected frequency of Unsafe behaviour in function of \(p_r^{\max}\). We observe that for such high $\beta$, even when CAS dominates, the expected frequency of Unsafe behaviour is very large, and it drops sharply to 0 when CS dominates ($Z=100,\beta=2,\mu=1/Z$).}
    \label{fig:si:strategy-frequencies}
\end{figure}

\begin{figure}[p]
    \centering

    \begin{subfigure}[t]{0.95\textwidth}
        \raggedright
        \textbf{A}\par\medskip
        \centering
        \includegraphics[width=\linewidth]{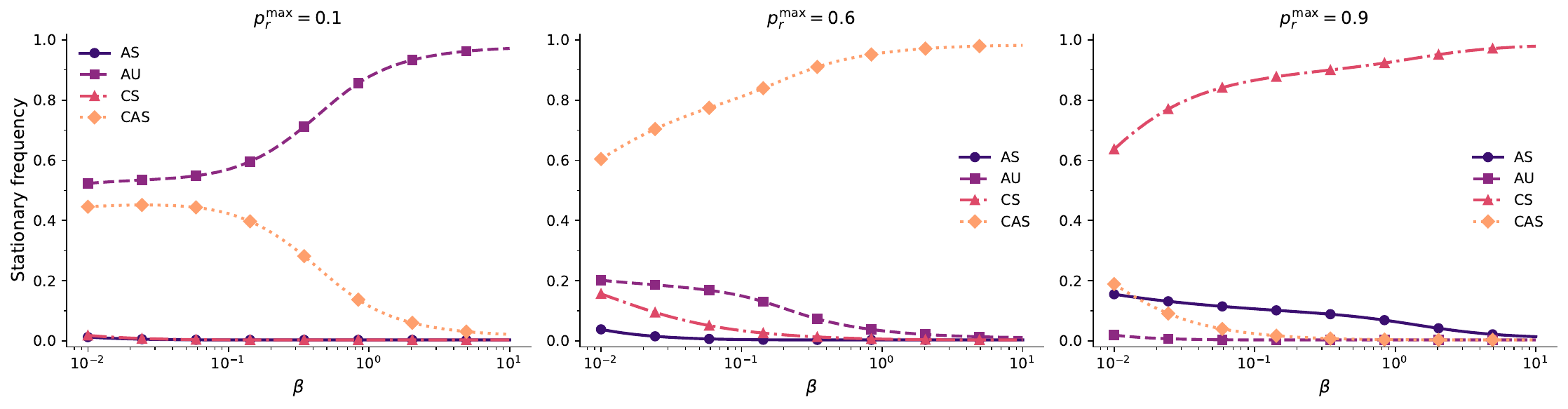}
        \label{fig:si:strategy-vs-beta}
    \end{subfigure}

    \vspace{1em}

    \begin{subfigure}[t]{0.95\textwidth}
        \raggedright
        \textbf{B}\par\medskip
        \centering
        \includegraphics[width=\linewidth]{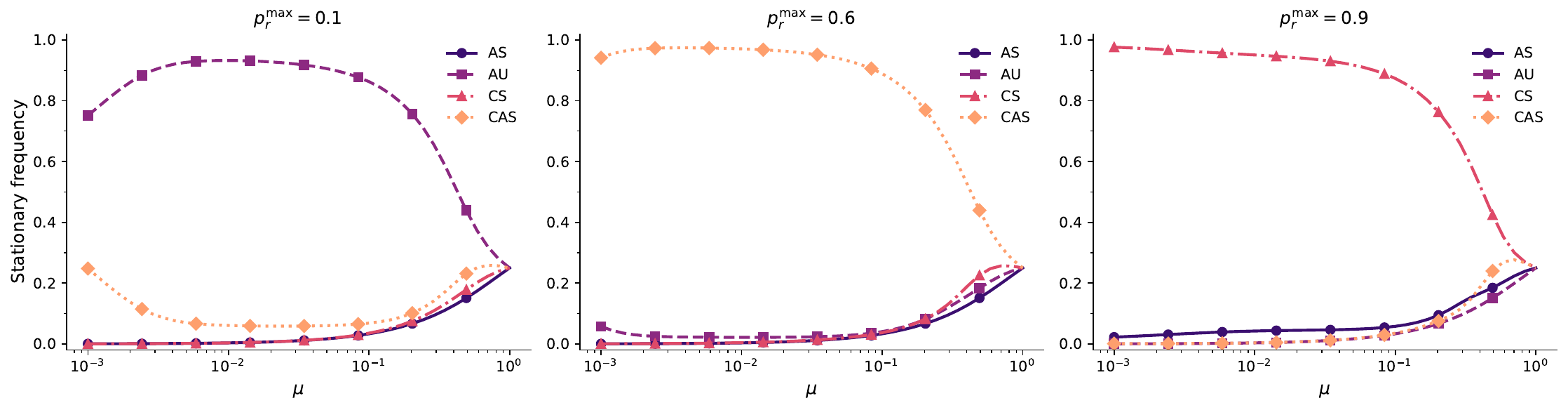}
        \label{fig:si:strategy-vs-mu}
    \end{subfigure}

    \caption{\textbf{Stationary frequency of strategies as a function of selection intensity and mutation rate.} (A) Stationary frequency of Always Safe (AS), Always Unsafe (AU), Conditionally Safe (CS), and Conditionally Antisocial Safe (CAS) as a function of the selection intensity \(\beta\) (log scale), for each \(p_r^{\max}\) treatment ($\mu=1/Z$).
    As \(\beta\) increases, the population concentrates increasingly on a single dominant strategy: AU at \(p_r^{\max}=0.1\), CAS at \(p_r^{\max}=0.6\), and CS at \(p_r^{\max}=0.9\), consistent with \Cref{fig:si:strategy-frequencies}.
    (B) Stationary frequency of the same four strategies as a function of the mutation rate \(\mu\) (log scale), for each \(p_r^{\max}\) treatment ($\beta=2$).
    As \(\mu\) increases, the stationary distribution becomes increasingly uniform across the four strategies; as \(\mu\) decreases, the same treatment-dependent ordering as in panel A is recovered (\(Z=100\)).}
    \label{fig:si:strategy-beta-mu}
\end{figure}

\Cref{fig:si:strategy-frequencies} shows how the predominance of each strategy (\Cref{fig:si:strategy-frequencies}A), and expected frequency of Unsafe actions (\Cref{fig:si:strategy-frequencies}B), is affected by the maximum private risk \(p_r^{\max}\). We observe that for low values of \(p_r^{\max}\), the Always Unsafe strategy (AU) dominates, while for higher values, the Conditionally Safe strategy (CS) becomes dominant. \Cref{fig:si:strategy-beta-mu} shows how the stationary distribution of strategies is affected by the selection intensity \(\beta\) and mutation rate \(\mu\). This analysis confirms that for \(\beta>1\) and for a large range of \(\mu\) the model results remain stable.

\subsection{Dynamics in the 3-strategy faces of the simplex}
\label{si:sec:dynamics-3strategy}

\begin{figure}[p]
    \centering
    \includegraphics[width=1.0\linewidth]{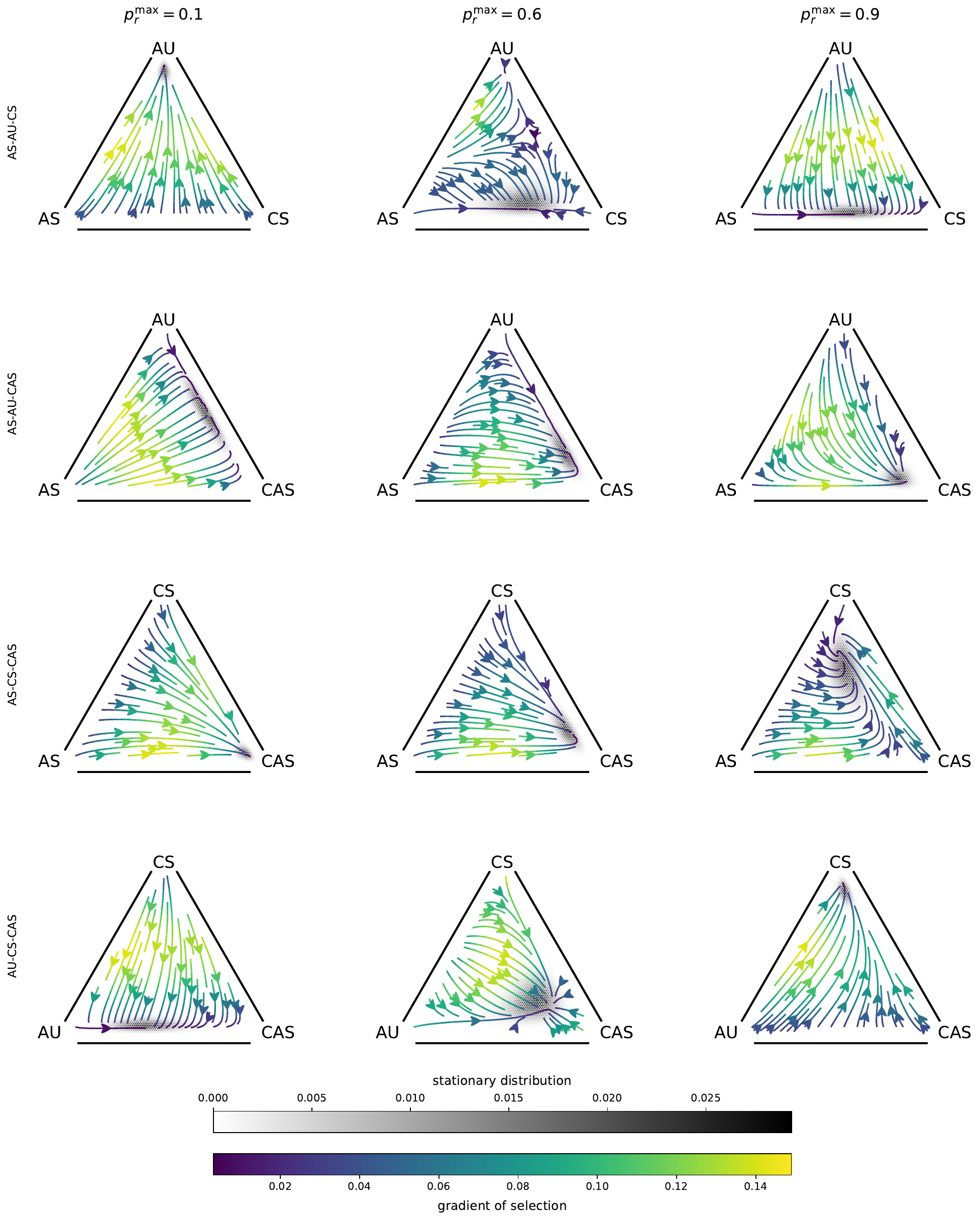}
    \caption{\textbf{Evolutionary dynamics in the four 3-strategy faces of the AS-AU-CS-CAS simplex.} Each row shows one of the four 3-strategy sub-games obtained by removing one strategy from the full model. Columns correspond to \(p_r^{\max}=0.1,0.6,0.9\). Streamlines show the direction of the selection gradient (colour encodes gradient intensity), and the grey-scale background shows the finite-population stationary distribution (\(Z=100,\beta=0.1,\mu=0.1\)).}
    \label{fig:si:strategy-dynamics-simplex}
\end{figure}

\Cref{fig:si:strategy-dynamics-simplex} plots the strategy dynamics in the four 3-strategy faces of the AS-AU-CS-CAS simplex. Each row shows one of the four 3-strategy sub-games obtained by removing one strategy from the full model (from top to bottom: AS-AU-CS, without CAS; AS-AU-CAS, without CS; AS-CS-CAS, without AU; AU-CS-CAS, without AS); columns correspond to \(p_r^{\max}=0.1,0.6,0.9\). As in \Cref{fig:model_predictions}C, streamlines show the direction of the selection gradient (colour encodes gradient intensity), and the grey-scale background shows the finite-population stationary distribution, with darker regions visited more frequently in the long run (\(Z=100,\beta=0.1,\mu=0.1\)). Across all four faces, stationary mass concentrates near AU at \(p_r^{\max}=0.1\), shifts towards CAS at \(p_r^{\max}=0.6\), and shifts towards CS at \(p_r^{\max}=0.9\), mirroring the treatment-dependent pattern observed in the full tetrahedral simplex (\Cref{fig:model_predictions}C). AS attracts negligible stationary mass in every face and at every treatment, confirming that AS is dominated by the conditional strategies regardless of which other strategy is present.

\end{document}